%% file: MAIN.tex
\documentclass[sigconf,screen]{acmart}

\usepackage{microtype}
\usepackage{balance}
\usepackage{xspace}
\usepackage{makecell}
\usepackage{adjustbox}
\usepackage{wrapfig}

\usepackage{xcolor}
\usepackage[position=bottom]{subcaption}
\usepackage{enumerate}
\usepackage{enumitem}
\usepackage{setspace}

\setlist[enumerate,1]{%
  label=\arabic*.,
}
\newlist{inlinelist}{enumerate*}{1}
\setlist*[inlinelist,1]{%
  label=(\roman*),
}

\usepackage{float}
\usepackage{listings}
\lstdefinestyle{customc}{
  numberstyle=\footnotesize,
  numbersep = 2pt,
  xleftmargin=5pt, %
  belowcaptionskip=1\baselineskip,
  breaklines=true,
  language=python,
  showstringspaces=false,
  abovecaptionskip=0pt,
  belowcaptionskip=1pt,
  basicstyle=\linespread{0.2}\footnotesize\ttfamily,
  keywordstyle=\bfseries\color{green!40!black},
  commentstyle=\itshape\color{purple!40!black},
  identifierstyle=\color{blue},
  stringstyle=\color{orange},
}
\newfloat{lstfloat}{t}{lop}
\floatname{lstfloat}{Listing}

\usepackage[ruled, vlined, linesnumbered, commentsnumbered, longend]{algorithm2e}
\usepackage{outlines}
\usepackage{makecell}

\setcopyright{acmcopyright}
\acmPrice{15.00}
\acmDOI{10.1145/3575693.3575707}
\acmYear{2023}
\copyrightyear{2023}
\acmSubmissionID{asplosb23main-p98-p}
\acmISBN{978-1-4503-9916-6/23/03}
\acmConference[ASPLOS '23]{Proceedings of the 28th ACM International Conference on Architectural Support for Programming Languages and Operating Systems, Volume 2}{March 25--29, 2023}{Vancouver, BC, Canada}
\acmBooktitle{Proceedings of the 28th ACM International Conference on Architectural Support for Programming Languages and Operating Systems, Volume 2 (ASPLOS '23), March 25--29, 2023, Vancouver, BC, Canada}
\received{2022-07-07}
\received[accepted]{2022-09-22}

\usepackage[normalem]{ulem}

\newcommand{\eg}{\textit{e.g.,}\xspace}
\newcommand{\ie}{\textit{i.e.,}\xspace}

\newcommand{\codelight}[1]{\underline{#1}}

\usepackage[nomessages]{fp}

\input{MACRO}

\begin{document}

\title{\sys: Generating Diverse and Valid Test Cases for Deep Learning Compilers} %

\author{Jiawei Liu}\authornote{Equal contribution.} %
    \affiliation{\institution{University of Illinois \\ at Urbana-Champaign}\city{Champaign}\state{IL}\country{USA}}
    \email{jiawei6@illinois.edu}
\author{Jinkun Lin}\authornotemark[1]
    \affiliation{\institution{New York University}\city{New York}\state{NY}\country{USA}}
    \email{jinkun.lin@nyu.edu}
\author{Fabian Ruffy}
    \affiliation{\institution{New York University}\city{New York}\state{NY}\country{USA}}
    \email{fruffy@nyu.edu}
\author{Cheng Tan}
    \affiliation{\institution{Northeastern University}\city{Boston}\state{MA}\country{USA}}
    \email{c.tan@northeastern.edu}
\author{Jinyang Li}
    \affiliation{\institution{New York University}\city{New York}\state{NY}\country{USA}}
    \email{jinyang@cs.nyu.edu}
\author{Aurojit Panda}
    \affiliation{\institution{New York University}\city{New York}\state{NY}\country{USA}}
    \email{apanda@cs.nyu.edu}
\author{Lingming Zhang}
    \affiliation{\institution{University of Illinois \\ at Urbana-Champaign}\city{Champaign}\state{IL}\country{USA}}
    \email{lingming@illinois.edu}

\begin{abstract}
Deep-learning (DL) compilers such as TVM and TensorRT are increasingly being used to optimize deep neural network (DNN) models to meet performance, resource utilization and other requirements. Bugs in these compilers can result in models whose semantics differ from the original ones, producing incorrect results that corrupt the correctness of downstream applications. However, finding bugs in these compilers is challenging due to their complexity. In this work, we propose a new fuzz testing approach for finding bugs in deep-learning compilers. Our core approach consists of (i) generating diverse yet valid DNN test models that can exercise a large part of the compiler's transformation logic using light-weight operator specifications; (ii) performing gradient-based search to find model inputs that avoid any floating-point exceptional values during model execution, reducing the chance of missed bugs or false alarms; and (iii) using differential testing to identify bugs. We implemented this approach in \sys which has found \bugTotal new bugs for \TVM, \TensorRT, \ORT, and PyTorch to date. Of these \bugConfirmed have been confirmed and \bugFixed have been fixed by their respective project maintainers.
\end{abstract}




\date{}
\maketitle

\input{sections/1-intro}

\input{sections/2-motivation}

\input{sections/3-overview}

\input{sections/3.1-abs-domain}

\input{sections/3.2-graph-gen}

\input{sections/3.3-param-fuzz}

\input{sections/3.4-inp-gen}

\input{sections/4-impl}
\input{sections/5-eval}

\input{sections/6-related-wk}

\input{sections/7-conclusion}

\input{ae}

\balance %
\bibliographystyle{ACM-Reference-Format}
\bibliography{reference}

\end{document}

%% file: MACRO.tex
\newcommand{\sys}{\textsc{NNSmith}\xspace}

\newcommand{\markStar}{\textcolor{orange}{$\bigstar$}\xspace}
\newcommand{\markOur}{\textcolor{orange}{$^\bigstar$}\xspace}
\newcommand{\PageOpt}[1]{}

\newcommand{\LEMON}{LEMON\xspace}
\newcommand{\GraphFuzz}{GraphFuzzer\xspace}
\newcommand{\FreeFuzz}{FreeFuzz\xspace}
\newcommand{\Tzer}{\textsc{Tzer}\xspace}

\newcommand{\fpev}{FP exceptional value\xspace}
\newcommand{\fpevs}{FP exceptional values\xspace}
\newcommand{\fpinf}{$Inf$\xspace}
\newcommand{\fpinfs}{$Inf$s\xspace}
\newcommand{\fpnan}{$NaN$\xspace}
\newcommand{\fpnans}{$NaN$s\xspace}

\newcommand{\TVM}{TVM\xspace}
\newcommand{\ORT}{ONNXRuntime\xspace}
\newcommand{\PyTorch}{PyTorch\xspace}
\newcommand{\TensorRT}{TensorRT\xspace}

\newcommand{\BC}{C_{b}} %
\newcommand{\attr}{\alpha} %

\newcommand{\invalidPercTwenty}{56.8\%\xspace}

\newcommand{\model}{M}
\newcommand{\layer}{F}
\newcommand{\loss}{\mathcal L}
\newcommand{\weight}{W}

\newcommand{\cons}{C}
\newcommand{\op}{\textit{op}}

\newcommand{\bugTotal}{72\xspace}
\newcommand{\bugConfirmed}{58\xspace}
\newcommand{\bugFixed}{51\xspace}

\newcommand{\bugIncon}{17\xspace}
\newcommand{\bugCrash}{55\xspace}

\newcommand{\bugTrn}{43\xspace}
\newcommand{\bugConv}{23\xspace}

\newcommand{\bugTorch}{10\xspace}

\newcommand{\bugTrnORT}{10\xspace}
\newcommand{\bugTrnTVM}{29\xspace}
\newcommand{\bugTrnTRT}{4\xspace}

\newcommand{\bugIntMismatch}{9\xspace} %

\newcommand{\bugLEMON}{17\xspace}
\newcommand{\bugGraphFuzz}{23\xspace}

\newcommand{\bugOurOnly}{49\xspace}
\newcommand{\bugOurOnlyTrn}{31\xspace}
\newcommand{\bugOurOnlyCvt}{14\xspace}

\newcommand{\bugSimpl}{5\xspace}
\newcommand{\bugSimplORT}{4\xspace}
\newcommand{\bugSimplTVM}{1\xspace}

\newcommand{\bugTrnFixed}{27\xspace}
\newcommand{\bugOptFixed}{21\xspace}
\newcommand{\bugTrnTVMLayoutExcl}{7\xspace}

\newcommand{\bugConvScalar}{7\xspace}
\newcommand{\bugConvBcast}{5\xspace}
\newcommand{\bugConvType}{7\xspace}

\newcommand{\covPercTVM}{18.6\%\xspace}
\newcommand{\covPercORT}{17.9\%\xspace}

\newcommand{\covAllHigherUniTVM}{10.8$\times$\xspace}

\newcommand{\covAllHigherUniORT}{32.7$\times$\xspace}

\newcommand{\covAllHigherFullTVM}{1.08$\times$\xspace}
\newcommand{\covPassHigherFullTVM}{1.09$\times$\xspace}
\newcommand{\covAllHigherFullORT}{1.8$\times$\xspace}
\newcommand{\covPassHigherFullORT}{1.85$\times$\xspace}

\newcommand{\covTzerHigher}{1.4$\times$\xspace}
\newcommand{\covTzerUniq}{461\xspace}

\newcommand{\LocCore}{5157\xspace}

\newcommand{\BinUniqRatioTVM}{1.8$\times$\xspace}

\newcommand{\BinUniqRatioORT}{2.2$\times$\xspace}
\newcommand{\BinTotalRatioORT}{2.3\%\xspace}

\newcommand{\BinOpRatio}{2.07$\times$\xspace}

\newcommand{\SuccRateTenBest}{98\%}

\newcommand{\cradle}{CRADLE\xspace}

\newcommand{\audee}{AUDEE\xspace}
\newcommand{\predoo}{Predoo\xspace}
\newcommand{\jsfun}{jsfunfuzz\xspace}
\newcommand{\tvmfuzz}{TVMFuzz\xspace}
\newcommand{\dl}{DL\xspace}
\newcommand{\csmith}{Csmith\xspace}

\newcommand{\deepsmith}{DeepSmith\xspace}
\newcommand{\deepfuzz}{DeepFuzz\xspace}

\newcommand{\langfuzz}{LangFuzz\xspace}
\newcommand{\montage}{Montage\xspace}
\newcommand{\gpt}{GPT\xspace}
\newcommand{\muffin}{Muffin\xspace}

%% file: sections/1-intro.tex
\section{Introduction}\label{sec:intro}
Deep learning (\dl) compilers such as \TVM~\cite{tvm}, \TensorRT~\cite{tensorrt}, and TensorFlow XLA~\cite{tensorflow} are increasingly being used to deploy deep neural network (DNN) models in many different applications.
These compilers optimize \dl models to meet desired performance, energy, and resource requirements, allowing their use by interactive or safety-critical applications deployed on a variety of devices. 
However, as compiler implementations are complex, we must be vigilant about detecting bugs in these systems.  Compiler bugs can result in crashes or generating an incorrect executable that produces different results than those intended by the user-specified input model\footnote{As deep-learning models use floating-point operations, a correctly compiled executable model can have close but not identical results as those of the input model. We do not regard this case as a bug.}. 

In this paper, we develop techniques to \emph{automatically find bugs in deep-learning compilers}.
Similar to prior work~\cite{tzer,luo2021graph,lemon}, we adopt a fuzzing and differential testing based approach: we generate random models, compile them using the compiler being tested, and then compare results obtained from the compiled model with those from a reference implementation. This basic approach faces two main challenges, which are not adequately addressed by prior work.  First, how do we generate structurally {\em diverse} and {\em valid} models?  Deep-learning compilers express a model as a computation graph of tensor operators.  For better test coverage, we must ensure model diversity, which requires us to generate graphs by combining operators in different ways.  However, connecting two arbitrary operators often produces invalid models, which are rejected by deep learning compilers. For example, a compiler will reject any computation graph containing a \emph{MatMul} (matrix multiplication) operator for which the number of rows in the first
input differs from the columns for the second. Therefore, for test efficiency, our graph generation method must also ensure the validity of generated models.  Second, given a compiled model, what weights/inputs should we use to run it for differential testing? 
Naively testing generated models with random or default weights/inputs can easily lead to floating point (FP) exceptional values, \ie \fpnans or infinities (\fpinfs). In such cases, we cannot compare the compiled model with its reference implementation. Therefore, to enable equivalence checking, we must be able to generate computation inputs that can avoid \fpev{}s during model execution.

We develop \sys, a tester for deep-learning compilers including \TVM~\cite{tvm}, \ORT~\cite{ort}, and \TensorRT~\cite{tensorrt}, that addresses these two challenges. 
\sys adopts a three-step approach for finding bugs: 
\begin{inlinelist}
    \item first, it automatically generates an arbitrary but valid computation graph expressing some model $\model_I$; 
    \item it then uses the compiler being tested to produce a compiled model $\model_O$ from $\model_I$, and a reference backend to produce an executable model $\model_R$; %
    and \item finally it generates random inputs 
    which it passes to $\model_O$ and $\model_R$, and compares their outputs.
\end{inlinelist}

\sys addresses the model and input generation challenges as follows.

\noindent\textbf{Generating diverse and valid computation graphs:} 
The computation graph expressing a deep-learning model consists of tensor operators with attributes attached to both the operators and graph edges. Operator attributes specify parameters such as kernel sizes that impact the operator's semantics, while edge attributes are used to specify input and output tensor types\footnote{A tensor's type defines its shape and its elements' data type.}. Before
proceeding with the actual compilation steps, deep-learning compilers check the validity of the input computation graph, \eg whether an operator's output tensor type matches the expected input tensor type of its downstream operators and whether an operator's attributes are valid. %
In order to produce valid graphs, \sys aims to capture and ensure the type matching constraints of a computation graph during its generation. To do so, \sys requires that users provide a specification for each operator, which specifies the constraints that must be satisfied by the operator's input tensors/attributes and also indicates the operator's output tensor type, which \sys can use to check the validity of generated graphs.
During \sys's incremental graph generation, it inserts one candidate operator at a time by solving for the satisfiability of its type matching constraints given the existing graph.  \sys uses an existing SMT solver~\cite{moura2008z3} for constraint solving.    %

\noindent\textbf{Executing compiled models without \fpevs.} In order to meaningfully compare the outputs of a compiled computation graph with those from a reference implementation, \sys aims to select computation inputs (aka model weights and inputs) that do not result in \fpnans or \fpinfs during execution.  Instead of random search, \sys uses gradient-guided search to efficiently find viable model inputs/weights for \SuccRateTenBest{} of the generated models with negligible overhead.

In addition to addressing the two main challenges above, %
we designed \sys so it can be easily extended to add support for new operators or to work with other deep-learning compilers. We do so  by providing users with a framework for writing operator specifications that are needed to ensure graph validity, and by providing a library of common patterns. In our experience, using this framework and library, users can write new operator specifications in a few lines of code. We evaluated the efficacy of our approach by using \sys to identify bugs in \TVM, \ORT, \TensorRT, and \PyTorch.

Over the last seven months, \sys found \bugTotal new bugs in these frameworks. Developers have confirmed \bugConfirmed and fixed \bugFixed of these bugs. Our coverage evaluation also shows that \sys outperforms the state-of-the-art fuzzer by \covAllHigherFullORT{} for \ORT and \covAllHigherFullTVM{} for \TVM in \textit{total} branch coverage, as well as \covAllHigherUniORT{} and \covAllHigherUniTVM{} respectively in \textit{unique} branches.

%% file: sections/2-motivation.tex
\section{Background}\label{sec:bg}

\subsection{The DNN Computation Graph}\label{sec:rwdnn}

\begin{figure}
    \centering
    \includegraphics[width=\linewidth]{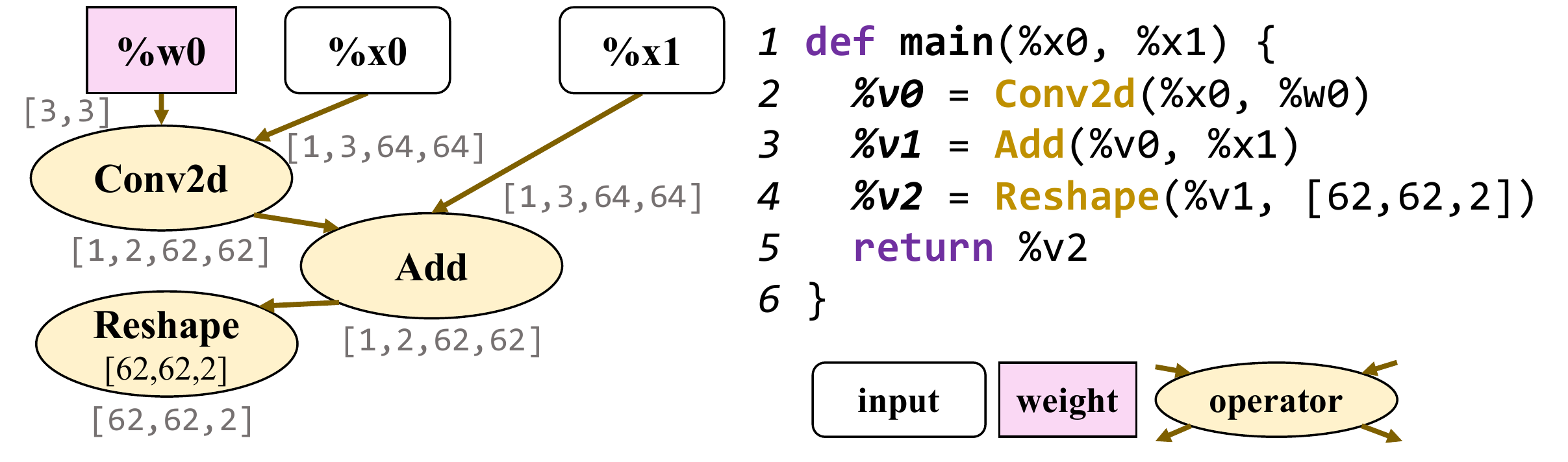}
    \caption{Sample DNN graph.}
    \label{fig:nnexp}
\end{figure}

\dl frameworks represent a model's underlying computation as a directed graph of tensor operators. In this work, we focus on DNN inference, where the graph captures the forward NN computation to generate predicted labels or outputs given the model weights and some inputs. For example, the model in Figure~\ref{fig:nnexp} is invoked by specifying its inputs (\ie input variables \texttt{\%x0} and \texttt{\%x1}) and the model weights (\ie input variable \texttt{\%w0}), and the DNN runtime computes the output tensor (\texttt{\%v2}) from these inputs.

In what follows, we use the term {\em tensor type} to refer to both the shape and element type of a tensor. In the DNN computation graph, each edge is marked with the tensor type that corresponds to the output of the edge's upstream operator, as shown in Figure~\ref{fig:nnexp}. When instantiating an operator, model developers must specify certain additional attributes that dictate its output tensor type. For example, on line 4 in Figure~\ref{fig:nnexp}, the \texttt{Reshape} operator takes \texttt{\%v1} as an input tensor and \texttt{[62,62,2]} as an attribute indicating the output shape.  Because each operator expects its input tensors to be of certain types, it is often invalid to connect two arbitrary operators together by an edge: \eg the reshape operator on line 4 is valid if and only if its upstream operator's output (\texttt{\%v1}) has $7688$ elements ($62\times 62\times 2$). This is akin to a ``type checking error'' in traditional programs. We say that a DNN computation graph is valid if and only if all operators in the graph are valid.

\subsection{DL Compilers}\label{sec:optdnn}

\begin{figure}[t]
    \centering
    \includegraphics[width=\linewidth]{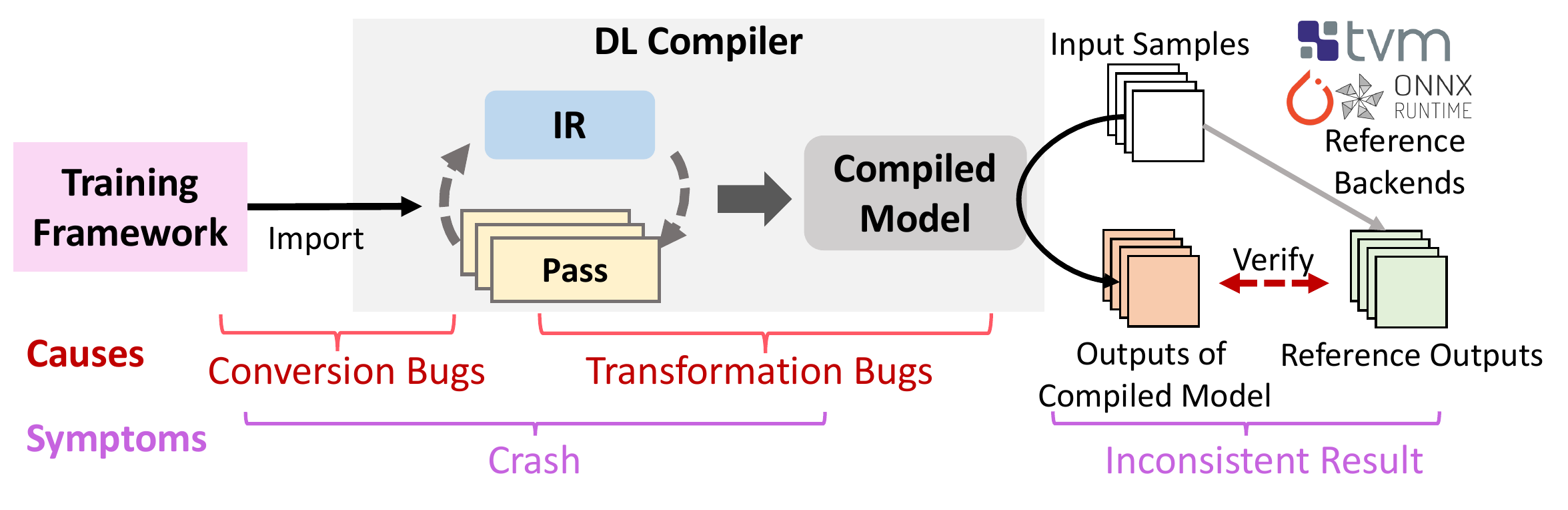}
    \caption{Deep learning compiler workflow and bug finding.}
    \label{fig:bugtarget}
\end{figure}

State-of-the-art DL compilers turn a user-specified model, expressed as a DNN computation graph, into an executable implementation.
As shown in Figure~\ref{fig:bugtarget}, DL compilers process an input DNN model in two stages during its compilation.

First, DL compilers need to convert an input computation graph into their own internal formats.  For interoperability, DL training frameworks typically export trained models 
to a standardized format such as ONNX~\cite{onnx}.
DL compilers take ONNX models as input and convert them to a compiler-specific Intermediate Representation (IR) that is easier for the compiler to ompimize.

Next, DL compilers invoke various transformation passes which rewrite the input IR into a more efficient version.
These passes include: graph-optimization passes that simplify the graph (\eg constant folding%
) or fuse operators (merge \emph{Add} and \emph{Softmax} into \emph{BiasSoftmax})~\cite{ortopt};
low-level passes that optimize computation using arithmetic simplification and loop tiling/fusion, to reduce computational overheads.

Compiler bugs can occur in both the conversion and transformation phases, but bugs in the later phase are generally harder to identify and debug. To comprehensively detect both kinds of bugs, we need to test using models with a diverse graph structure and tensor operators.

\subsection{Challenges in Finding DL Compiler Bugs}\label{subsec:bug-ex}
Differential testing and fuzzing~\cite{mckeeman1998differential} is a promising approach for finding DL compiler bugs. As shown in the right part of Figure~\ref{fig:bugtarget}, this approach requires synthesizing random models for compilation, then running the compiled models with random inputs, and finally comparing the generated results with those from a reference implementation. %

\PageOpt{
Such bugs manifest themselves in two ways (\emph{symptoms} in Figure~\ref{fig:bugtarget}) i) compiler internal errors for valid models; and ii) inconsistent results when comparing with reference outputs from other backends.
Furthermore, compiler bugs can attributes to various causes and we focus on \emph{causes} shown in Figure~\ref{fig:bugtarget}, esp. transformation bugs.
}

There are several challenges facing the basic approach of fuzzing and differential testing, which are not addressed by prior work~\cite{luo2021graph, lemon, wei2022free}. Next, we illustrate these challenges using concrete examples.

\begin{lstfloat}
\begin{lstlisting}[language=Python, caption={DNN patterns that can expose compiler bugs.\label{lst:runningex}}]
def M0():            # M0 triggers a compiler crash bug!
  A = Conv2d(...)    # shape: (1,2,1,48)
  B = Ones(1,1,48)   # shape: (  1,1,48)
  return A + B

def M1():@\label{lst:runningex:slice}@            # bug NOT triggered!
  A = Conv2d(...)    # shape: (1,2,1,48)
  B = Ones(1,2,1,@\codelight{49}@) # different shape: (1,2,1,49)
  return A + B[:,:,:,@\codelight{:48}@] # slice to match (1,2,1,48)

def M2():            # bug NOT triggered!
  A = Conv2d(...)    # shape: (1,2,1,48)
  B = Ones(@\codelight{1,1,1}@)     # trivial shape: (1,1,1) @\label{lst:runningex:1b}@
  return A + B

def M3(): @\label{lst:runningex:layoutbug}@           # M3 can trigger a semantic bug
  Y = Conv2d(Conv2d(...), ...)  # bug lies here
  Y = @\codelight{Pow(}Y@, BIG_NUM@\codelight{)}@ @\label{lst:runningex:nantensor}@# bug not exposed due to Infs
  return Y
\end{lstlisting}
\end{lstfloat}

\noindent\textbf{Challenge \#1: Generating graphs with diverse patterns.} Finding DL compiler bugs  requires generating input graphs that contain a variety of operators and connections. Some prior fuzzers~\cite{wei2022free} test only using single-operators and thus are too limiting. \LEMON~\cite{lemon} and \GraphFuzz~\cite{luo2021graph} generate multi-operator computation graphs, but they are restricted to certain types of operators and connections in order to avoid ``type check'' errors on the generated graphs (detailed in \S\ref{sec:rwmlsystest}).  These restrictions limit graph diversity, and compromise test coverage.  

Listing~\ref{lst:runningex} shows an example model ({\tt M0}) generated by \sys which has triggered a layout analysis bug in \TVM.  \LEMON cannot generate this\PageOpt{ example} model because {\tt M0} contains non-shape-preserving operators (\eg
\emph{Conv2d}) and connections (\eg broadcasting) which are not supported by \LEMON
for ensuring graph validity.
\GraphFuzz uses a different strategy to guarantee graph validity. Specifically, \GraphFuzz tries to ``fix'' mismatched tensor shapes in generated graphs through slicing and padding, as illustrated by line~\ref{lst:runningex:slice} in model {\tt M1}. Unfortunately, doing so biases the generated graphs to include many slicing/padding nodes. In our example, the slice operation in {\tt M1} would silence the layout bug found in {\tt M0}. 

\noindent\textbf{Challenge \#2: Exploring diverse attributes for operators and edges.}  When generating graphs, it is tempting to ignore the need to explore the operator/edge attribute space and rely on some default values.
For example, {\tt M2} of Listing~\ref{lst:runningex} uses trivial attributes (\eg always 1) to initialize operator \texttt{Ones(1,1,1)} (line~\ref{lst:runningex:1b}).  Unfortunately, the bug found by {\tt M0} will not be triggered by {\tt M2}.  Since exploring different attribute values results in diverse output tensor types on edges, it further complicates the task to ensure the validity of generated graphs.

\noindent\textbf{Challenge \#3: Running compiled models to produce numerically valid output.}  Using arbitrary inputs and model weights to test a compiled model can result in \fpevs (\ie \fpnans and \fpinfs) during execution for differential testing.   Such cases occur when the given inputs to some operator are outside of its expected domain, \eg feeding \emph{Sqrt} negative values results in \fpnans, and feeding $\emph{Pow}$ large base or exponents results in \fpinfs. Larger graphs are particularly prone to encountering \fpevs. For example, we have found that \fpnan/\fpinf occurs in \invalidPercTwenty of 20-node models generated by \sys when using PyTorch's default weight initializer. Previous testing frameworks did not consider these issues, and consequently up to 41\% of their bug reports can be false-alarms because of the undefined/non-deterministic behaviors arising from the \fpnan/\fpinf~\cite{graphfuzzbug}.

Clearly we should not compare the output of a compiled model to those of the reference implementation if the results themselves contain \fpnan/\fpinf. 
What about those scenarios with normal final results (aka without any \fpnan/\fpinf) where some internal operator has produced \fpevs during graph execution? For example, operator {\tt ArgMax} can output a normal FP value even though one of its upstream operators gives it \fpnan as input.
It is a subtle requirement that we must also exclude these results from differential testing or risk incurring false positives in bug detection. This is because when handling \fpevs, otherwise semantically equivalent operators could produce different results.  Therefore, to be able to test effectively, we must generate model inputs/weights that avoid \fpevs for all operators in the graph. Only then we refer to the model's output as {\em numerically valid}.  Otherwise, we might miss detecting bugs. As an example, Listing~\ref{lst:runningex}'s model {\tt M3} can trigger a semantic bug. However, this bug is not exposed because the execution results in \fpinf values which are not used for comparison.

%% file: sections/3-overview.tex
\section{\sys's Design}\label{sec:tech}

\begin{figure*}
    \centering
    \includegraphics[width=\linewidth]{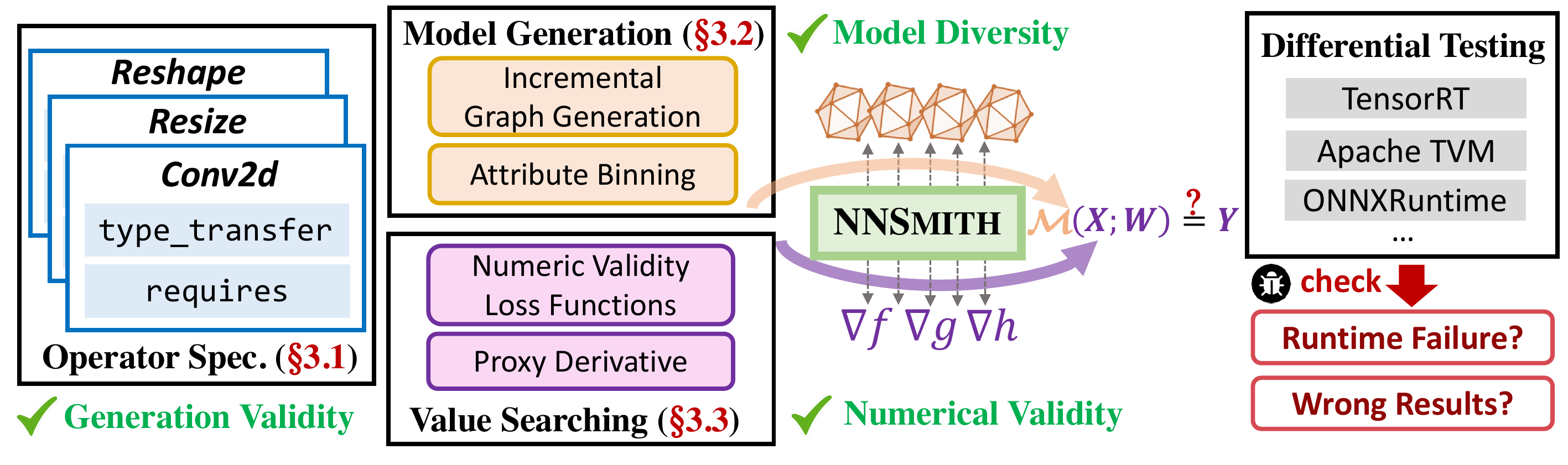}
    \caption{Overview of \sys.}
    \label{fig:overview}
\end{figure*}

\paragraph{Overview of the approach.} Figure~\ref{fig:overview} shows an overview of \sys's workflow. \sys generates random models that are valid, which are then compiled and executed. \sys takes as input a compiler's type checking requirements in the form of operator specifications (\S\ref{sec:model}), and then uses an SMT solver to generate graphs and operator attributes that meet these constraints (\S\ref{sec:graphgen}). Next, when running a compiled model, \sys uses a gradient guided search procedure to find benign weights/inputs so that no \fpevs are produced at any step of the execution (\S\ref{sec:grad}). Finally, \sys compares the results obtained from multiple deep learning libraries and compilers to those from a reference implementation to identify bugs.

%% file: sections/3.1-abs-domain.tex
\subsection{Modeling DNN Operators}
\label{sec:model}

\sys generates random DNN models expressed as computational graphs by connecting together different operators. We aim to generate valid graphs that ``type check'', \ie graphs where each operator's attributes and input tensor type meet requirements imposed by the compiler.

In order to generate valid graphs, we require users to provide operator specifications that explicitly state the compiler's requirements for each operator and guarantees about its output. An operator's specification codifies rules for checking validity and  depends on its inputs and attributes: For example, the 2-D convolution operator (\emph{Conv2d}) has several attributes, including a kernel, and takes an image as input. A model that uses a \emph{Conv2d} operator is valid if the input image is a rank-4 tensor that is larger than the kernel's size. 

\begin{lstfloat}
\begin{lstlisting}[language=Python, caption={Sample specification for the 2D Pooling operator\label{lst:pool}.}]
class Pool2d(AbsOpBase): @\label{lst:pool:cls}@
  input_type =   # [(input#1 types), ...]
      [(AbsTensor(float32, 4), AbsTensor(float64, 4))]  @\label{lst:pool:in}@
  output_type =  # [(output#1 types), ...]
      [(AbsTensor(float32, 4), AbsTensor(float64, 4))] @\label{lst:pool:out}@
  
  def __init__(self, kh, kw, stride, pad): @\label{lst:pool:init}@
    self.kh, self.kw = kh, kw ... @\label{lst:pool:members}@

  def requires(self, inputs :List[AbsTensor]): @\label{lst:pool:requires}@
    ih, iw = inputs[0].shape[-2:]
    return [self.kw > 0, self.kh > 0, @\label{lst:pool:consexp}@
      self.stride > 0, self.pad >= 0, 
      self.kw <= 2 * self.pad + iw, ...] 

  def type_transfer(self, inputs :List[AbsTensor]): @\label{lst:pool:typeinf}@
    n, c, iw, ih = inputs[0].shape
    oshape = [n, c,
      (h - self.kh + 2*self.pad) // self.stride + 1,
      (w - self.kw + 2*self.pad) // self.stride + 1]
    return [AbsTensor(inputs[0].dtype, shape=oshape)] @\label{lst:pool:typeinfret}@
    
  def infer_input_type(self, outputs):@\label{lst:pool:inpinf}@# @\S\ref{sec:graphgen}@
    return [AbsTensor(outputs[0].dtype, 4)]
\end{lstlisting}
\end{lstfloat}

While our implementation includes specifications for common operators (detailed in \S\ref{sec:impl}), we designed \sys so that it is easy for users to write specification for additional operators. \sys specifications are written using symbolic integers and abstract tensors. An abstract tensor is specified by its data type, rank and shape. In our implementation we specify an abstrat tensor's data type and rank using concrete values, and use symbolic integers to specify its shape. As we will see later in \S\ref{sec:graphgen}, \sys uses an SMT solver to assign concrete integers to each symbolic integer during graph generation. \sys operator specifications provide input and output types (specified using abstract tensors), constraints on inputs and attributes, as well as transfer rules for each operator. Listing~\ref{lst:pool} shows the operator specification for a 2-D pooling operator (\emph{Pool2d}), and we describe each of part below:

\noindent\textbf{Inputs and outputs.} An operator's attributes are inferred from the inputs to its \texttt{\_\_init\_\_} function. The class variables \texttt{input\_type} and \texttt{output\_type} describe the input and output tensor types respectively (Lines~\ref{lst:pool:in} and~\ref{lst:pool:out}). Programmers specify a list of tuples, each tuple says what data types can be used for an input (or provided as output). In the listing, the \emph{Pool2d} operator accepts a single rank 4 tensor of 32-bit or 64-bit floats.

\noindent\textbf{Constraints.}\label{sec:cons} The operator's \texttt{requires} function (Line~\ref{lst:pool:requires}) returns constraints that its inputs and attributes must satisfy as a list of logical predicates. For example, among other constraints, the \emph{Pool2d} operator requires that the kernel size should be greater than 0 (Line~\ref{lst:pool:consexp}).

\noindent\textbf{Type transfer function.} The operator uses a \emph{type transfer function} (Line~\ref{lst:pool:typeinf}) to specify how its output tensor relates to its inputs. For example, on Line~\ref{lst:pool:typeinfret}, \emph{Pool2d}'s type transfer function relates the shape of the operator's output tensor to its kernel size (\texttt{self.kw} and \texttt{self.kh}) and its input shapes. Observe that the constraints output by the type transfer function become the input constraints on a downstream operator. These constraints are used to to combine constraints from connected operators in a computation graph, thereby allowing \sys to generate valid models.

%% file: sections/3.2-graph-gen.tex
\subsection{Model Generation}\label{sec:graphgen}
Given a set of operator specifications, \sys generates models that are \emph{topologically diverse} and whose operators use \emph{diverse attributes}. Below we first detail our approach for generating diverse model topologies and then present our binning based approach to assigning diverse attributes.

\noindent\textbf{Generating computation graphs.}
Our model generation algorithm is designed to ensure that generated computation graphs are fully connected, as is the case with most real-world models. Additionally, it is also designed so that it can generate a rich variety of models, including ones similar to existing multi-modal and multi-task models~\cite{he2017mask,baltruvsaitis2018multimodal,radford2021learning,ngiam2011multimodal} that can accept multiple inputs and/or produce multiple outputs.

\sys generates connected computation graph by extending an existing graph while maintaining connectivity. It does so by starting with a graph that contains a single \emph{placeholder} node, and extending it by either (a) adding a new node whose input edges are connected to the output of an existing node (we refer to this as a \emph{forward} insertion) or (b) replacing an existing placeholder node with an operator node whose input edges are connected to one or more placeholder nodes (we refer to this as \emph{backward} insertion). In both cases, the node added by \sys is picked at random from the set of symbolic operator specifications ($\op$) it is provided.  Placeholder nodes have one output, and at the end of the graph generation process they are replaced by input nodes or by weights (which are constant inputs). Algorithm~\ref{algo:binsert} shows our graph generation algorithm. We detail the steps taken when inserting a randomly selected operator ($\op$) into an existing compute graph below:

\begin{enumerate}[leftmargin=*]
\item \emph{Type matching:} To insert $\op$, \sys must first find a feasible insertion point in the current graph. When using forward insertion, this means finding an output edge in the graph whose constraints (as provided by the operator that node represents) satisfy $\op$'s input constraints. Similarly, when using backward insertion, this means finding a placeholder node whose output is connected to node(s) whose input constraints are satisfied by $\op$'s output constraints. To do so we need to check constraint satisfaction, and we use a SMT solver for this. Rather than invoking an SMT solver for all possible insertion points, we use a simple type matching heuristic to filter out nodes that are obviously infeasible because of incompatible data types or ranks. For example, when using forward insertion for \emph{Where(cond, T, F)}, type matching (Lines~\ref{algo:finsert:typematch}) will filter out any output edges which are not boolean.
\item \emph{Constraint solving:} Next, \sys generates constraints for any feasible insertion points that have not been filtered out by its type matching heuristic, and uses an SMT solver to check their satisfiability. \sys caches constraints for the current model (in $\model.$\texttt{solver}) to reduce constraint generation overheads, and uses incremental solving~\cite{bjorner2018programming} to reduce time taken for checking constraints (Line~\ref{algo:binsert:trysolve}).

\item \emph{Node insertion:} As we stated previously, we use one of two approaches to insert nodes into the graph: forward insertion and backward insertion:
    \begin{itemize}
        \item \emph{Forward insertion} (Line~\ref{algo:binsert:forward}) selects one group of plausible tensors ($v$) as the inputs of $\op$ (Line~\ref{algo:binsert:ichoose}) and inserts $\op$ as their consumer (Line~\ref{algo:binsert:finsert}) if the insertion constraints are satisfiable (Line~\ref{algo:binsert:fsolve}).
        
        \item \emph{Backward insertion} (Line~\ref{algo:binsert:backward}) replaces an existing placeholder node with $\op$. To do so it first determines a placeholder candidate ($v$) by matching $\op$'s output type and the candidate's type (Line~\ref{algo:binsert:typematch}---\ref{algo:binsert:pchoose}). Next, it infers $\op$'s input type from $v$. If $\op$'s input type constraints can be satisfied for the candidate, \sys replaces the candidate with $\op$ and creates new placeholder nodes (of the inferred types) to act as $\op$'s inputs (Lines~\ref{algo:binsert:binsert} and~\ref{algo:binsert:breplace}).
    \end{itemize}
\end{enumerate}

\begin{algorithm}
\small
\caption{Computation graph generation.}\label{algo:binsert}
\DontPrintSemicolon
\SetKwProg{Fn}{Function}{:}{}
\SetKw{Continue}{continue}\SetKw{Break}{break}\SetKw{Raise}{raise}

\SetKwInOut{Input}{Input}
\Input{Graph $\model$; operator to try $\op$; global constraints $\cons$.}

\SetKwFunction{ForwardInsert}{\textsc{ForwardInsert}}
\SetKwFunction{BackwardInsert}{\textsc{BackwardInsert}}
\SetKwFunction{TypeMatch}{\textsc{TypeMatch}}
\SetKwFunction{Solve}{\textsc{Solve}}

\SetInd{0em}{1.13em} %

\Fn{\Solve{$\model$, $\cons$, $\op$, $\op$'s inputs $v$}}{\label{algo:binsert:solve}
$\cons\leftarrow \cons \cup op.$\texttt{requires}($v$)\;\label{algo:binsert:ccbeg}
\For{$\forall s \in op.\texttt{type\_transfer}(v)$}{
    $\cons\leftarrow\cons \cup \{s\mathrm{.shape}_0 \ge 1,\cdots,s\mathrm{.shape}_{s.rank-1} \ge 1\}$\;
}\label{algo:binsert:ccend}
\Return $\model$\texttt{.solver.try\_add\_constraints}($\cons$)\;\label{algo:binsert:trysolve}
}

\Fn{\ForwardInsert{$\model$, $\op$}}{\label{algo:binsert:forward}
$S\leftarrow$ \TypeMatch{$\model.$\texttt{intermediates()}, $\op$.\texttt{input\_type}}\;\label{algo:finsert:typematch}
$v\leftarrow$ randomly choose one \emph{input} combination from $S$\;\label{algo:binsert:ichoose}
\If{\Solve{$\model$, $\emptyset$, $\op$, $v$}}{\label{algo:binsert:fsolve}
    $\model$.\texttt{insert\_consumer}(v, $\op$)\;\label{algo:binsert:finsert}
}
}

\newcommand\mycommfont[1]{\footnotesize\ttfamily\textcolor{blue}{#1}}
\SetCommentSty{mycommfont}

\Fn{\BackwardInsert{$\model$, $\op$}}{\label{algo:binsert:backward}
$S\leftarrow$ \TypeMatch{$\model.$\texttt{placeholders()}, $\op$.\texttt{output\_type}}\;\label{algo:binsert:typematch}
$v\leftarrow$ randomly choose one set of \emph{placeholder}s from $S$\;\label{algo:binsert:pchoose}
\tcp{Also see \texttt{infer\_input\_type} in Line~\ref{lst:pool:inpinf} of Listing~\ref{lst:pool}}
$p\leftarrow$ new placeholders w.r.t. $\op.$\texttt{infer\_input\_type}(v)\;\label{algo:binsert:pinps}
$o\leftarrow \op\texttt{.type\_transfer}(p)$\;\label{algo:binsert:ttrans}
\If{\Solve{$\model$, $\{\forall i\in[0,|v|], v_i\mathrm{.shape} = o_i\mathrm{.shape}\}$, $\op$, $p$}}{\label{algo:binsert:bsolve}
    $\model$.\texttt{insert\_consumer}(p, $\op$)\;\label{algo:binsert:binsert}
    $\model$.\texttt{replace\_placeholder}(v, $\op$)\;\label{algo:binsert:breplace}
}
}
\end{algorithm}

%% file: sections/3.3-param-fuzz.tex
\noindent\textbf{Attribute binning.}\label{sec:paramfuzz}
In addition to topological diversity, attribute diversity is also crucial as discussed in \S\ref{subsec:bug-ex}. We use the solution (model) generated by an SMT solver (Z3~\cite{moura2008z3} in our implementation) when checking satisfiability for the graph's constraints to determine attributes. However, we found that the models produced by SMT solvers tend to pick boundary values for integer constraints.
For example, when a tensor shape $D$ is constrained so that $\{d_i\ge 1; d_i \in D\}$, the Z3 SMT solver always returns models where $\{d_i=1;d_i \in D\}$, limiting the diversity of attributes in generated DNNs.

We address this problem by adding extra constraints (which we call binning constraints) that limit each attribute to a randomly chosen range. Algorithm~\ref{algo:binning} shows how we generate these binning constraints: We start from an empty set of binning constraints ($\BC$ in Line~\ref{algo:binning:emptyset}), and iterate over each attribute $\attr$ of each operator $\op$ (Line~\ref{algo:binning:oloop}-\ref{algo:binning:iloop}) in the graph ($\model$). Note that this algorithm also considers placeholders as operators, and uses placeholder tensor shape as attributes for these operators.
The binning constraints for each attribute are generated by randomly choosing from one of $k$ bins, where the $i^{th}$ bin represents the range $[2^{i-1}, 2^i)$ (when $i < k$, the last bin represents the range $[2^{k-1},\infty)$), and limiting the attribute to a subset of the bin's range. To do so, we randomly pick a bin (Line~\ref{algo:binning:selectbin}) for each $(\op, \attr)$, and sample two integers (\ie $l$ and $r$) from it (Line~\ref{algo:binning:callgbin}).
We then add $l \le \attr \le r$ as the binning constraints for the $(\op, \attr)$ pair.
We use bins with exponential ranges (Line~\ref{algo:binning:samplefrombin}) because, in practice, systems are more sensitive to changes in smaller values, e.g., changing a variable from 0 to 1 generally has larger effect on the output than changes from 30 to 31. %
Our approach of dividing ranges in exponential buckets is inspired by how AFL~\cite{afl} coarsely records the hit counts for execution tuples~\cite{aflbucket}.
We allow operators to provide a different, more specialized strategy, for attribute binning, and in this case we simply use the provided constraints ($\cons^*$ in Line~\ref{algo:binning:specialize}) instead (details in \S\ref{sec:impl}).

Adding extra binning constraints to the graph's constraints can produce an unsatisfiable constraint system, leading to a situation where we can find no attributes for a valid graph. 
We avoid this situation by adding constraints \textit{only} after a graph has been generated (Paragraph 2 in \S\ref{sec:graphgen}) and \textit{only} when doing so does not impact satisfiability.
Specifically, if the solver fails to find a satisfiable assignment after adding $\BC$ to the graph's constraints, we randomly drop half of the constraints and retry, until it succeeds (Line~\ref{algo:binning:drop}).

\begin{algorithm}
\small
\SetInd{0em}{1.13em}
\caption{Attribute binning.}\label{algo:binning}

\newcommand\mycommfont[1]{\footnotesize\ttfamily\textcolor{blue}{#1}}
\SetCommentSty{mycommfont}

\DontPrintSemicolon
\SetKwProg{Fn}{Function}{:}{}
\SetKwInOut{Input}{Input}
\Input{Graph $\model$; global constraints $\cons$; \# bins $k$. %
}

\SetKwFunction{SampleFromBin}{\textsc{SampleFromBin}}
\SetKwFunction{AttrBinning}{\textsc{AttrBinning}}

\Fn{\SampleFromBin{$i$, $k$}}{\label{algo:binning:samplefrombin}
\eIf{$i \neq k$}{
    $b, t \sim \mathrm{Uniform}(i-1, i)$ where $ b < t $ and $ b,t\in \mathbb R$\; %
    \Return ($\lfloor 2^b \rfloor, \lfloor 2^t \rfloor$)\;
}{
    \Return $(2^{k-1},\infty)$\;
}
}

\Fn{\AttrBinning{$M$, $C$, $k$}}{\label{algo:binning:attrbin}

\texttt{$\BC\leftarrow\emptyset$} \tcp*{Binning constraints to apply}\label{algo:binning:emptyset}
\For(){$\op\in\model$\label{algo:binning:oloop}}{
    \For(){$\attr\in\op$\label{algo:binning:iloop}}{ %
        \eIf{no specialization for $(\op,\attr)$}{
            $i \gets$ random integer $\in[1,k]$\tcp*{Randomly pick a bin}\label{algo:binning:selectbin}
            $l, r \gets$ \SampleFromBin{$i$, $k$}\;\label{algo:binning:callgbin}
            $\BC\leftarrow \BC \cup \{l \le \attr \le r\}$\;\label{algo:binning:econs} %
        }{
            $\BC\leftarrow \BC \cup \cons^*(\op,\attr)$\tcp*{The last paragraph of \S\ref{sec:impl}}\label{algo:binning:specialize}
        }
    }
}

\While{$\neg\model$\texttt{.solver.try\_add\_constraints($\BC$)}}{
  $\BC\leftarrow$ randomly select $\frac{|\BC|}{2}$ constraints in $\BC$ \label{algo:binning:drop}
}

}
\SetInd{0em}{0.75em} %
\end{algorithm}

%% file: sections/3.4-inp-gen.tex
\subsection{Improving Numeric Validity with Gradients}\label{sec:grad}

Next, \sys generates inputs and weights that can be used to test the generated models. We initially considered using randomly selected numbers. However, we found that the generated graphs produce \fpevs, including \emph{NaN} (not a number) and \emph{Inf} (infinite number).
For example, when generating 20-operator graphs, \fpevs occur in \invalidPercTwenty{} of generated graphs if we use random weights and inputs.%

This is because some operators, which we refer to as \emph{vulnerable operators}~\cite{gradientnumeric}, produce real (\eg $\sqrt{x}$ returns \fpnan if $x < 0$) or stable (\eg $x^y$ returns \fpinf for large $x$ and $y$) results only for a subset of their input domain. If a vulnerable operator's input lies outside of this domain, the operator outputs an \fpev, which propagates through the model and impacts the model's output, preventing us from comparing model outputs during differential testing. Table~\ref{tab:vulop} lists examples of vulnerable operators we encountered in our evaluation.

\begin{table}
\adjustbox{max width=\linewidth}{
    \centering
    \begin{tabular}{c c c c}
    \hline
    \textbf{Operator} & \textbf{Domain} & \textbf{Violation} & \textbf{Loss functions}\\
    \hline
    \emph{Asin}(X)   & $|X|\le 1$ & \emph{NaN} & 
    $\loss(|X|-1\leq 0)$\\
    
    \emph{Div}(X, Y) & $|Y|>0$ & \emph{NaN} & 
    $\loss(|Y| > 0)$\\
    
    \emph{Pow}(X, Y)         & $\displaystyle \left\{ {X>0 \atop Y\log(X) \leq 40} \right.$  & \emph{NaN}/\emph{Inf} & 
    $\displaystyle \left\{ {\loss(X > 0) \atop \loss(Y \log(X) - 40 \leq 0)} \right.$ \\
    
    \emph{Log2}(X)    & $X>0$ & \emph{NaN} & 
    $ \loss(X > 0)$ \\
    \hline
    \end{tabular}
}
    \caption{Representative vulnerable operators.}
    \label{tab:vulop}
\end{table}

\begin{table}
    \centering
    \begin{tabular}{c c}
     \hline
    \textbf{Tensor Ineq.} & \textbf{Loss function $\loss$} \\
    \hline
    $f(X) \leq 0$ & $\sum_{x\in X} \max(f(x), 0)$\\
    $f(X) <    0$ & $\sum_{x\in X}\max(f(x)+\epsilon, 0)$\\
    \hline
   \end{tabular}
   \caption{Tensor inequality to loss function conversions.}
   \label{tab:ineq-loss}
\end{table}

One way to address this problem is to use additional heuristics to extend and fix vulnerable operators. For example, changing \emph{Div}(x, y) to \emph{Div}(x, |y|+$\epsilon$) ensures that the \emph{Div} operator is safe. However, this requires changing operator inputs, which limits graph diversity as discussed in \S\ref{subsec:bug-ex}.
We thus propose an alternate approach, where we use a gradient-search algorithm to find inputs that ensure that the model's output is numerically valid.
Our approach is inspired by GRIST~\cite{gradientnumeric}, though that work has the \emph{opposite} goal: it aims to find inputs that result in \fpevs.

At a high-level, our approach associates a set of loss functions with each operator. When selecting inputs, \sys starts with random inputs, and then iteratively refines these inputs so that no operator in the graph produces an \fpev. In each iteration, \sys identifies the first operator in the model that produces an \fpev. It then uses the set of loss functions associated with the operators to compute new model inputs and uses these for the next iteration. The algorithm terminates when no \fpevs are found. We provide details below:

\noindent\textbf{Loss functions for avoiding \fpevs.} \sys associates a set of loss functions with each operator, which our input search algorithm (Algorithm~\ref{algo:grad}) uses to update inputs to avoid \fpevs. Users can specify loss functions for each operator, and below we describe our approach to producing loss functions.

As we noted above, vulnerable operators produce valid outputs (\ie outputs that are not \fpevs) when inputs are drawn from a particular domain, and this domain can be expressed (or approximated) by the conjunction of a few (usually one or two) inequality predicates on the operator's input. We refer to this conjunction of inequality predicates as the operator's \emph{tensor inequalities}. For example, the $Sqrt(X)$ operator takes a tensor $X$ as input, and is numerically valid if and only if $X\geq 0$ (\ie all elements of $X$ are positive). Similarly, the $Pow(X, Y)$ operator's numerically valid domain can be under-approximated as $X\geq 0 \wedge Y\log(X) \leq 40$, which requires that all elements of $X$ be positive to avoid \fpnans (since $Y$ might contain fractional elements) and bounds $Y\log(X)$ to avoid outputs that are too large (and would be represented by infinity)~\footnote{We constrain the logarithm instead of directly using the power function to ensure that the loss function does not generate a \fpev.}. We associate a loss function with each predicate in an operator's tensor inequality. We do so by first rewriting each predicate so that it is either of the form $f(X) < 0$ or $f(X) \leq 0$, and then use the formulas in Table~\ref{tab:ineq-loss} to convert this cannonical form to a scalar loss. We show examples of the loss functions produced in this manner in Table~\ref{tab:vulop}. When an operator produces invalid outputs, the search algorithm picks which loss function to use by finding a predicate that is violated by the operator's current input and using the loss function associated with it. For simplicity, our design assumes that a loss function is positive if and only if its associated predicate is violated by the operator's input, allowing us to use any positive loss function associated with the operator (Line~\ref{algo:grad:loss}) without evaluating its associated predicate.

\noindent\textbf{Proxy derivative.} Given a vulnerable operator's loss, \sys uses gradient propagation to compute changes to the model inputs and weights. Doing so requires computing gradients (derivatives) for each operator in the graph (Line~\ref{algo:grad:optx}). However, some operators are either undifferentiable for some inputs (\eg \emph{Floor}, \emph{Ceil}, and other operators cannot be differentiated at integers)  or have zero gradient in some region (\eg \emph{ReLU} has gradient 0 for all negative inputs), and this prevents backward propagation. For these functions, we use \emph{Proxy Derivative Functions}~\cite{Bengio2013EstimatingOP} instead of actual derivatives during gradient propagation.

Given an operator $\layer$ whose gradient is $0$ in region $\mathbb U$ we use $\frac{d \layer(x)}{d x}(x\in \mathbb U) = \alpha$ as the derivative. We set $\alpha$'s sign based on the overall trend of the function, \eg we use a positive $\alpha$ for \emph{ReLU} because it is monotonic. Similar to \emph{LeakyReLU}~\cite{xu2015empirical}, we choose a small magnitude for $\alpha$, thus avoiding large discrepancies between the proxy and the actual derivative. On the other hand, if the operator $\layer$ cannot be differentiated in the region $U$, we use the closest left-derivative instead.

\newcommand{\LabelName}{\textbf{\textcolor{orange}{\textbf{\texttt{OUTER}}}}}
\SetInd{0em}{1.13em}

\begin{algorithm}[h!]
\small
\caption{Gradient-guided value search.}\label{algo:grad}
\DontPrintSemicolon
\SetKwProg{Fn}{Function}{:}{}
\SetKw{Continue}{continue}
\SetKw{Break}{break}
\SetKw{Raise}{raise}

\newcommand\mycommfont[1]{\footnotesize\ttfamily\textcolor{blue}{#1}}
\SetCommentSty{mycommfont}

\SetKwFunction{GradSearch}{\textsc{GradSearch}}
\Fn{\GradSearch{DNN $\model$, learning rate $\mu$}}{
$\langle X,\weight\rangle  \leftarrow$ randomly initialized inputs and weights\;\label{algo:grad:init}
\LabelName{:}\While{time budget not exhausted} {\label{algo:grad:while}
    \For{operator $\layer_i$ in \texttt{TopologicalSort}($\model$) }{\label{algo:grad:topo}
        $I_i\leftarrow$ input to $\layer_i$\;
        $O_i\leftarrow \layer_i(I_i)$\;
        \If{$\exists$ NaN/Inf $\in O_i$}{\label{algo:grad:numtest}
            $\loss\leftarrow$ first positive loss functions of $\layer_i$\;\label{algo:grad:loss}
            $\langle X,\weight\rangle\leftarrow \langle X,\weight\rangle - \mu \nabla_{\langle X,\weight\rangle}\loss(I_i)$\;\label{algo:grad:optx}
            \If(\tcp*[h]{Zero gradients}){$\langle X,\weight\rangle \text{ not changed}$}{
                $\langle X,\weight\rangle\leftarrow$ randomly initialized values\;\label{algo:grad:reinit}
            }
            \ElseIf{$\exists$ NaN/Inf $\in \langle X, \weight \rangle$}{
                Replace NaN/Inf with random values\;\label{algo:grad:reinit2}
            }
            \Continue \LabelName\tcp*[l]{Go to Line~\ref{algo:grad:while}} \label{algo:grad:brk}
        }
    }
        \Return $\langle X,\weight\rangle$\;
}
\Raise failed to find viable $\langle X,\weight\rangle$\;\label{algo:grad:fail}
}
\end{algorithm}

\noindent\textbf{Search process.} The overall input search algorithm (Algorithm~\ref{algo:grad}) proceeds as follows: Given a model $\model$ and time budget $T$, we first randomly initialize inputs and weights $\langle X, \weight\rangle$ (Line~\ref{algo:grad:init}) used by the first iteration of the search algorithm (Line~\ref{algo:grad:while}). In each iteration, we find the first operator (in topological order, Line~\ref{algo:grad:topo}) that produces an \fpev (Line~\ref{algo:grad:numtest}). We use its loss function as an optimization objective (Line~\ref{algo:grad:loss}) to tune $\langle X, \weight\rangle$. If the gradient is neither zero nor a \fpevs then we move on to the next iteration (Line~\ref{algo:grad:brk}), otherwise we restart the search with a different initial value (Line~\ref{algo:grad:reinit} and~\ref{algo:grad:reinit2}). The algorithm throws an exception (Line~\ref{algo:grad:fail}) if it does not terminate within the time budget. 

Because loss functions can vary by orders-of-magnitude across operators, we use Adam~\cite{adam}, an adaptive learning rate scheduling algorithm, to set the learning rate. We also reset the learning rate whenever we switch the loss functions used for optimization (as would be the case when an iteration finds a different operator). While this design can lead to a scenario where optimizing for one operator leads to another producing invalid outputs and vice-versa, we found that this to be rare in practice (it occurred less than 1\% of the time). We found that the most common reason for the search algorithm failing was that the model has no valid inputs.

%% file: sections/4-impl.tex
\section{Implementation}\label{sec:impl}

\sys is implemented in \LocCore{} lines of Python code.
Consistent with Algorithm~\ref{algo:binsert}, \PageOpt{within certain graph size and timeout budgets, }\sys outputs a symbolic graph and its SMT solution for being valid with the help of the Z3~\cite{moura2008z3} solver.
We then concretize the symbolic graph by invoking the materialized PyTorch functors
in the topological order,
and export the model to the deployment-friendly ONNX~\cite{onnx} format using PyTorch's exporter.
We also use PyTorch to implement our algorithm for finding model inputs/weights that result in numerically valid output (\S\ref{sec:grad}).

Since DL compilers vary in operator and data type support, we infer the set of operators supported by the compiler being tested by trying to compile single-operator models with different data types. We use this information when generating graphs, so as to avoid ``\emph{Not-Implemented}'' errors.

When verifying outputs from the compiled model, we regard \PyTorch's results as the oracle.
We use \PyTorch as the reference backend over compiler cross-checking because:
1) Obtaining results from \PyTorch is a ``\emph{free}'' lunch which is a by-product of gradient-based value searching; %
2) Current DL compilers support different operator sets or data types, implying that cross-checking is limited to the common set of their support matrices; and
3) \PyTorch's results are more trustworthy for being a better-tested interpreter, which has been used to produce oracles in many downstream compilers.
In rare
cases value inconsistency (1 out of \bugTotal{} in our bug finding) can happen in \PyTorch's converter which makes it unclear whether the bug comes from the converter or the compiler. Therefore, for better fault localization, if the compiler and \PyTorch disagree on the model outputs, we further compare it with the compiler ``O0'' mode at an extra cost of re-compilation.
If the ``O0'' mode also disagrees with the optimized model, we then are confident the compiler's optimization must be wrong.

We wrote operator specifications in \sys using information obtained from framework documentation~\cite{torch11conv} and source code~\cite{ortshapeinfer}. To simplify this task, we implemented several meta types including, unary/binary, reduce and broadcast that further reduce the amount of code needed to specify an operator. Using these, we found that we could implement 59 (out of 73) operator specification within 4 lines of code. Furthermore, even for the most complex specification, which was for \emph{Conv2d}, the \texttt{requires} function has 9 inequalities and the \texttt{type\_transfer} function is only 7 lines of code (formatted by PEP8~\cite{pepeight}) that can be quickly implemented in a few minutes.
Furthermore, these specifications can be written once and then shared by all compilers that can accept ONNX models as input.

Regarding the $\cons^*$ in attribute binning (Line~\ref{algo:binning:specialize} in Algorithm~\ref{algo:binning}), our current implementation uses the following default settings: 1) we add one extra bin that contains only one integer ``0'' for the ``padding'' attribute in \emph{Conv2d}, for that padding can also be 0; 
2) similarly, for the ``padding'' attribute in \emph{ConstPad}, \emph{ReplicatePad}, and \emph{ReflectPad}, we add both the 0-bin and negative bins that for supporting zero and negative padding; 
3) we specifically handle for the indexing ranges (\ie ``start'' and ``end'' attributes) in \emph{Slice} to make sure the range is valid to its input tensor shape.

%% file: sections/5-eval.tex
\section{Evaluation}\label{sec:eval}

\subsection{Experimental Setup}\label{sec:setup}

\noindent\textbf{Metrics.}
We mainly target the following metrics for evaluation:
\begin{itemize}[leftmargin=*]
    \item \emph{Code coverage}: Following prior fuzzing work~\cite{bohme2022reliability,bohme2017directed,wei2022free}, we trace source-level branch coverage for both the entire systems and their pass-only components, measuring 1) \emph{total} coverage counts all hit branches; and 2) \emph{unique} coverage counts unique branches (``hard'' branches) that other baselines cannot cover.
    \item \emph{Bug counting}:
    Following prior work~\cite{weimer2006patches,wei2022free,tzer}, we use the number of independent patches as the number of detected bugs, except that we directly count the number of bug reports for closed-source systems (\ie \TensorRT) and unfixed ones.
    \PageOpt{For example, in \TVM, 6 types of \emph{Reduce} operators (\eg max. reduce) fail when the input is scalar.
    We only count it as one bug as simply fixing code in the parent \emph{Reduce} class can resolve all such issues.
    Meanwhile, a few similar cases happen in other meta classes (\eg \emph{Squeeze}), which require other independent patches.
    We thus distinguish them as different bugs.}
\end{itemize}

\noindent\textbf{Baselines.}
We compare \sys with both the state-of-the-art general DNN model generators (\LEMON and \GraphFuzz) and fuzzer specifically designed for \TVM (\ie \Tzer).

\begin{itemize}[leftmargin=*]
    \item \LEMON~\cite{lemon} is a mutation-based model generator that mutates pre-trained Keras~\cite{keras} models~\cite{lemon_website}. We convert Keras models into ONNX, to reuse the same differential testing and evaluation framework of \sys for fair comparison;
    \item \GraphFuzz~\cite{luo2021graph} generates models by randomly connecting nodes from a block corpus. While \LEMON is limited to shape-preserving unary operators, \GraphFuzz also supports non-unary operators by aligning input tensor shapes with slicing/padding and uses specific attributes to create shape-preserving instances for a few non-shape-preserving operators such as \emph{Conv2d}. As its implementation is not open-sourced, for a fair comparison, we reimplemented its main design, \eg stitching operators via padding/slicing, by replacing \sys's specification-based node insertion.
    \item \Tzer~\cite{tzer} is a coverage-guided and mutation-based fuzzer targeting \TVM's low-level IR. As DNNs generated by \sys can also be lowered to low-level IR, we compare \Tzer with \sys to see if \sys can well cover low-level optimizations as \Tzer. 
\end{itemize}

\noindent\textbf{Systems under test.}
\sys finds bugs in the following commonly used compilers:

\begin{itemize}[leftmargin=*]
    \item \ORT~\cite{ortopt} (by Microsoft) is a graph-optimized DNN library for ONNX models, with over 130 source files on various graph optimizations. Like many runtime-based frameworks (\eg PyTorch), though \ORT enables optimizations, the optimized graph will still be directly mapped into pre-compiled kernel functions (\ie no code generation). To evaluate pass-only coverage, only files under \texttt{onnxruntime/core/optimizer} are instrumented;
    \item \TVM~\cite{tvm} is an end-to-end compiler for deploying DNNs on various platforms. In addition to 61 graph-level passes, \TVM also performs up to 58 low-level optimizations to generate highly optimized target code.
    As a front end, ONNX models will be converted into \TVM's graph-level IR to perform further optimization.
    \TVM also has a much higher coverage upper limit (\ie 116k) than \ORT (\ie 65k) given its higher capability/complexity.
    For pass-only instrumentation, we consider files in all \texttt{transfroms} folders.
    \item \TensorRT~\cite{trt-pop} is a compiler and runtime highly optimized for NVIDIA GPUs and has been used by more than 350k developers across 27.5k companies.
    Since \TensorRT is closed-sourced, we exclude it for coverage evaluation.
\end{itemize}

\noindent\textbf{Experimental configuration.}
The testbed hardware configurations include:
1) Intel 10700k CPU (16 threads); 2) 64 GB memory (3200 Mhz); and 3) 2TB NVMe SSD.
The operating system is Ubuntu 20.04 and targeted DL systems are compiled by Clang 14 under release mode.
Except that we performed bug findings on various latest compiler versions over the last tenvspace months, the default software versions used in evaluation are: \ORT v1.12 (\texttt{c556f5}), \TVM v0.8 (\texttt{9ab3a1}), TensorRT v8.4 and \PyTorch v1.13 (\texttt{dev20220615}).

When evaluating \sys, for Algorithm~\ref{algo:binsert} we choose between forward and backward at every insertion randomly with equal probability.
For the binning approach we use $k=7$ bins (\S\ref{sec:graphgen}) to ensure a decent amount of attribute diversity while keeping the models small for fuzzing efficiency.
For the gradient search, the initial learning rate is set to be 0.5, $\epsilon$ in the tensor inequality loss function is set to $10^{-10}$.
While \LEMON does not explicitly control the graph sizes (since it mutates existing models), we set the default generated graph size of \sys and \GraphFuzz to be 10.
For coverage evaluation, we run fuzzers for 4 hours by default (following \Tzer~\cite{tzer}) as we observe that code coverage curves generally converge before that point (\eg as shown in Figure~\ref{fig:totalcov}).

\subsection{End-to-end Coverage Efficiency}\label{sec:eval:cov}

\begin{figure}
\centering
\begin{subfigure}{0.5\linewidth}
    \includegraphics[width=\linewidth]{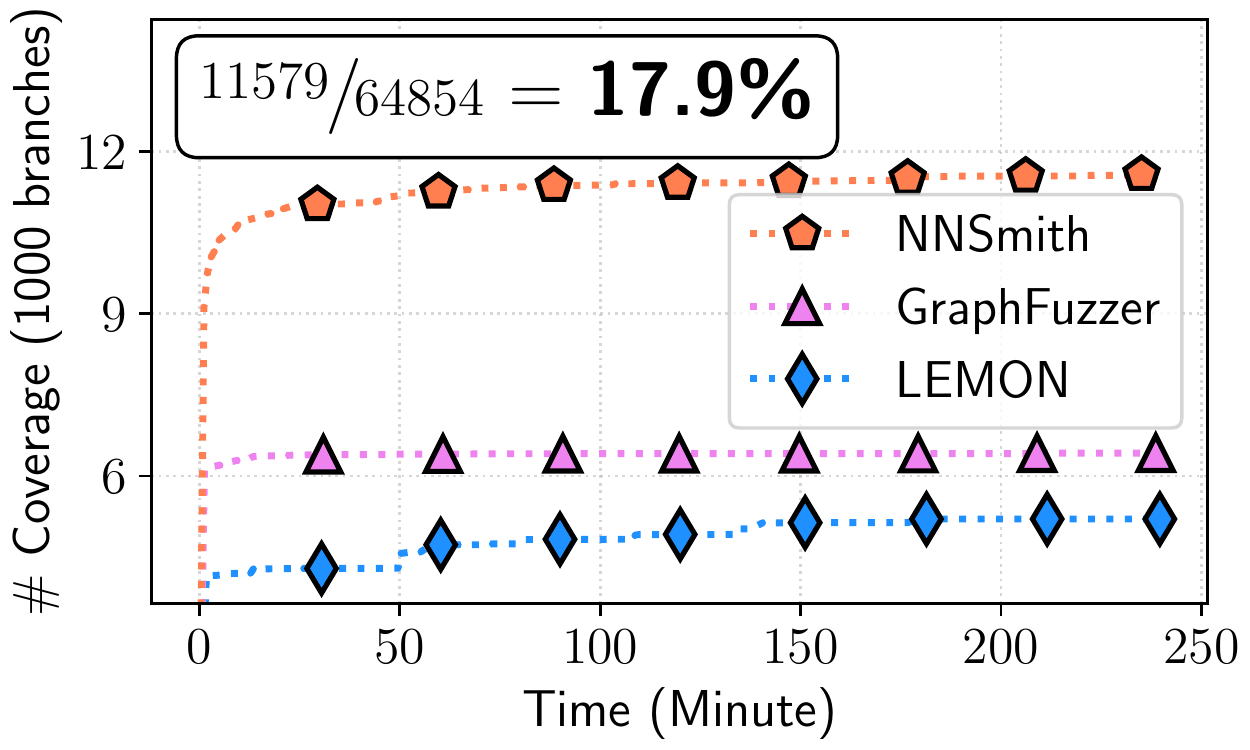}
    \caption{\ORT}
    \label{fig:orttotal}
\end{subfigure}
\hspace{-1.5mm}
\begin{subfigure}{0.5\linewidth}
    \includegraphics[width=\linewidth]{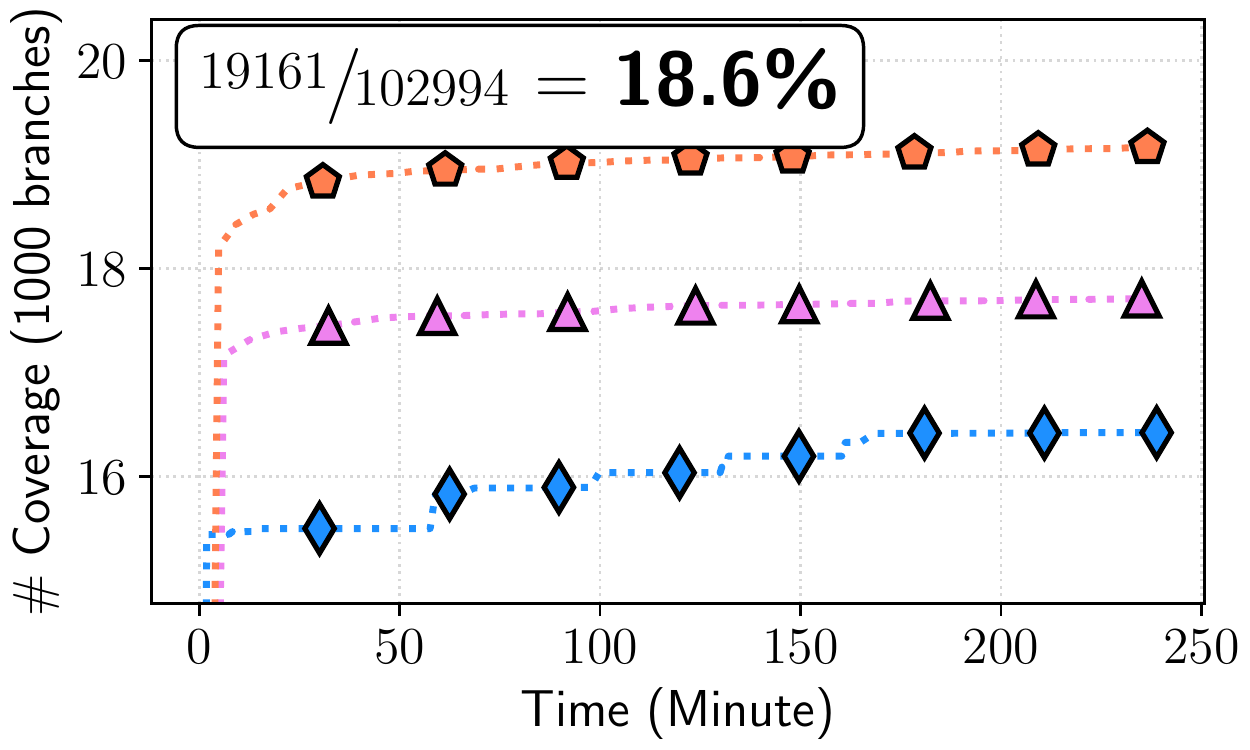}
    \caption{\TVM}
    \label{fig:tvmtotal}
\end{subfigure}
\caption{Total branch coverage over time (\emph{all} files).}
\label{fig:totalcov}
\end{figure}

\begin{figure}
\centering
\begin{subfigure}{0.5\linewidth}
    \includegraphics[width=\linewidth]{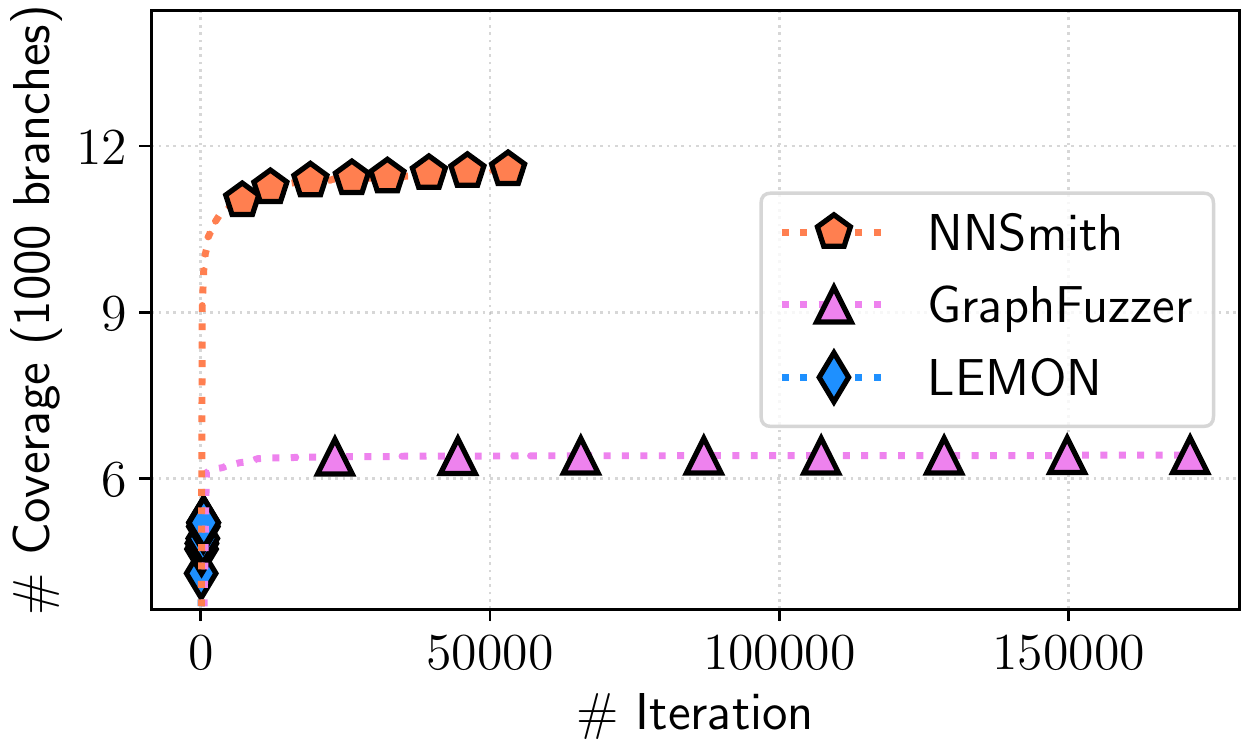}
    \caption{\ORT}
    \label{fig:orttotal:iter}
\end{subfigure}
\hspace{-1.5mm}
\begin{subfigure}{0.5\linewidth}
    \includegraphics[width=\linewidth]{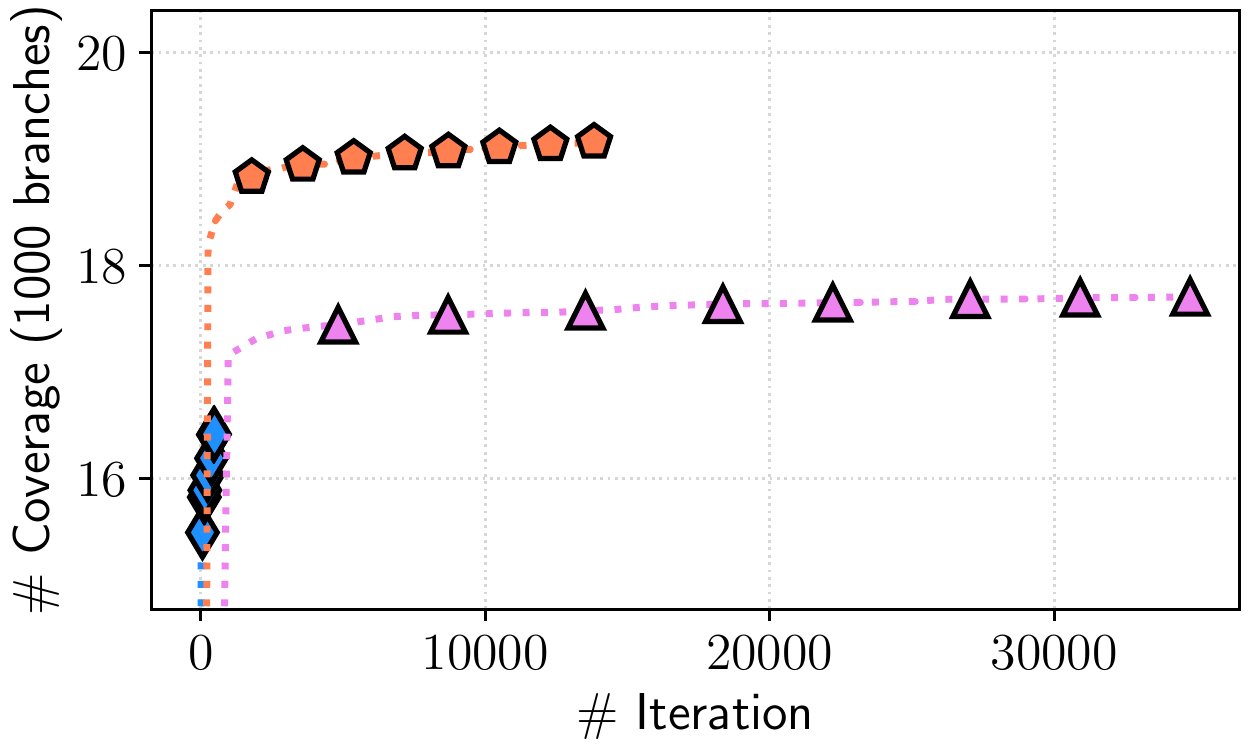}
    \caption{\TVM}
    \label{fig:tvmtotal:iter}
\end{subfigure}
\caption{Total branch coverage over test cases (\emph{all} files).}
\label{fig:totalcov:iter}
\end{figure}

\begin{figure}
\centering
\begin{subfigure}{0.5\linewidth}
    \includegraphics[width=\linewidth]{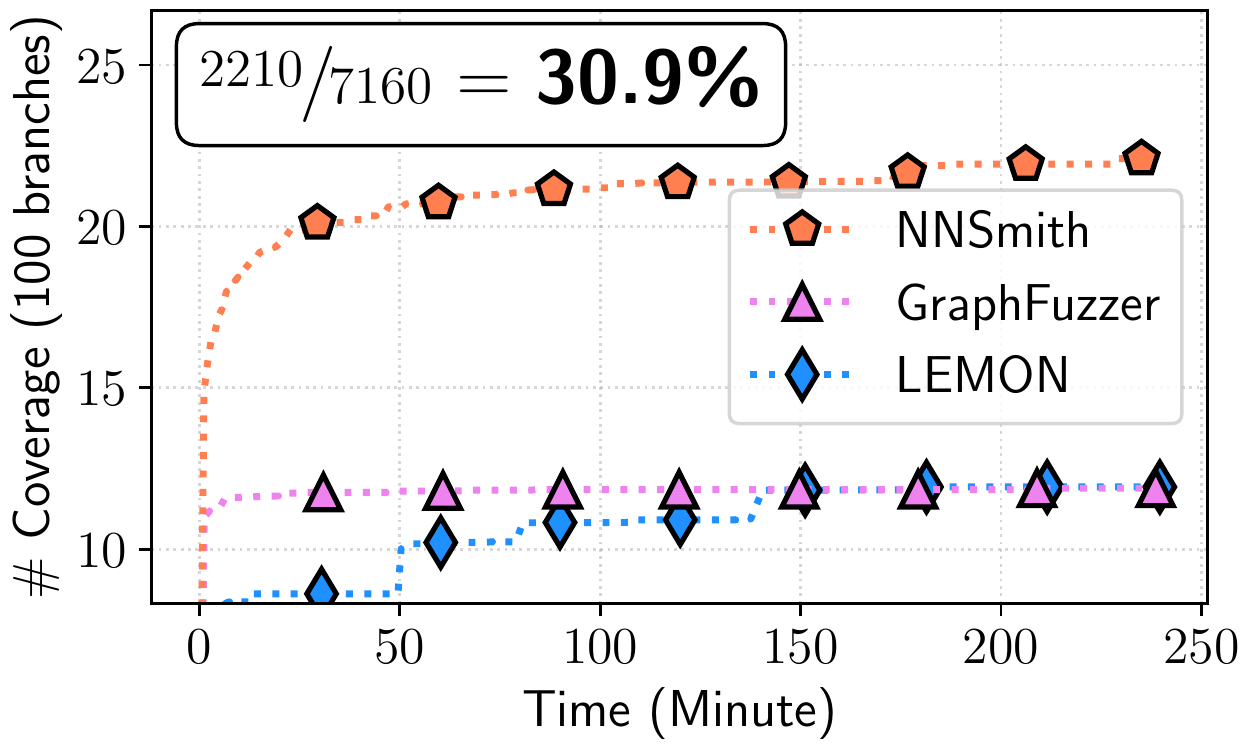}
    \caption{\ORT}
    \label{fig:ortpass}
\end{subfigure}
\hspace{-1.5mm}
\begin{subfigure}{0.5\linewidth}
    \includegraphics[width=\linewidth]{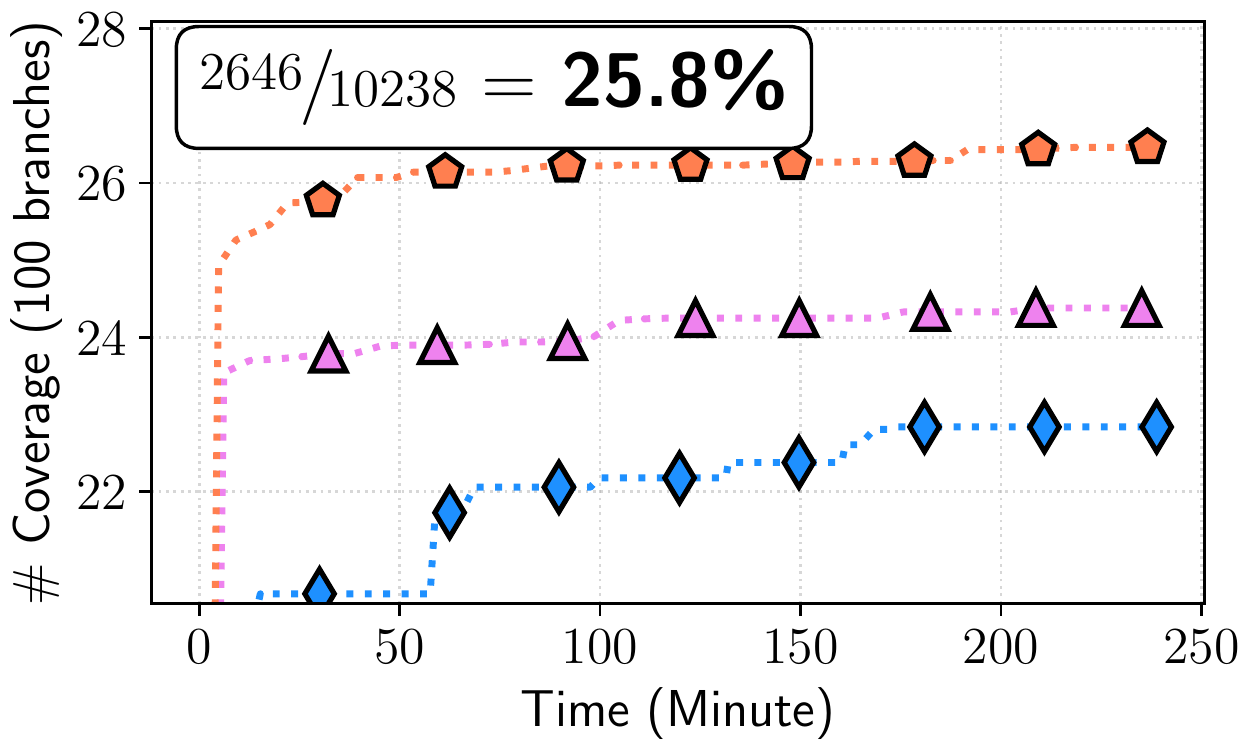}
    \caption{\TVM}
    \label{fig:tvmpass}
\end{subfigure}
\caption{Total branch coverage over time (\emph{pass} files).}
\label{fig:passcov}
\end{figure}

\begin{figure}
\centering
\begin{subfigure}{0.49\linewidth}
    \includegraphics[width=\linewidth]{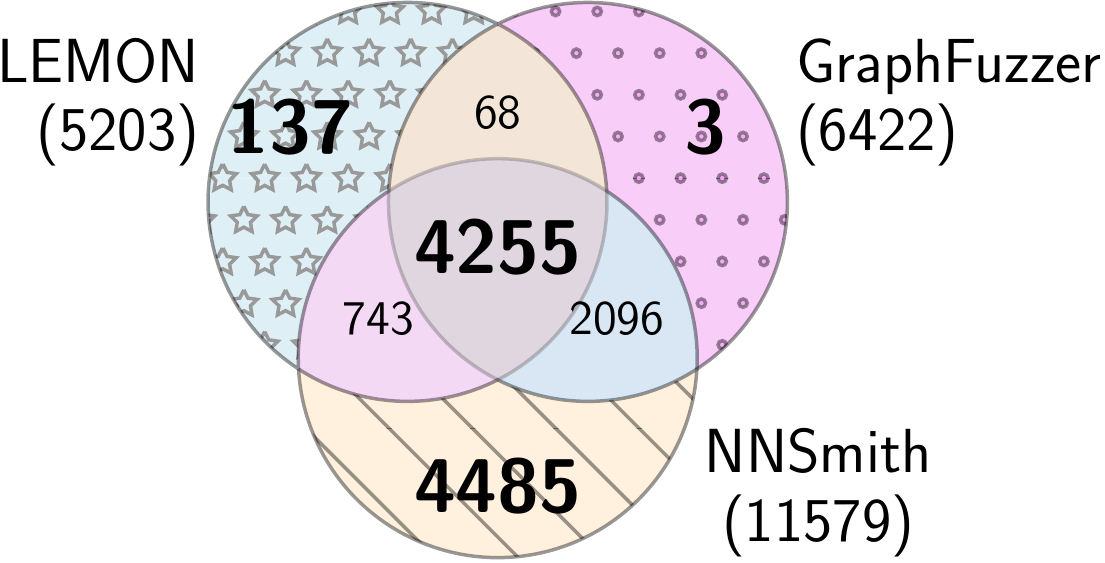}
    \caption{\ORT}
    \label{fig:ort_venn:all}
\end{subfigure}
\hfill
\begin{subfigure}{0.49\linewidth}
    \includegraphics[width=\linewidth]{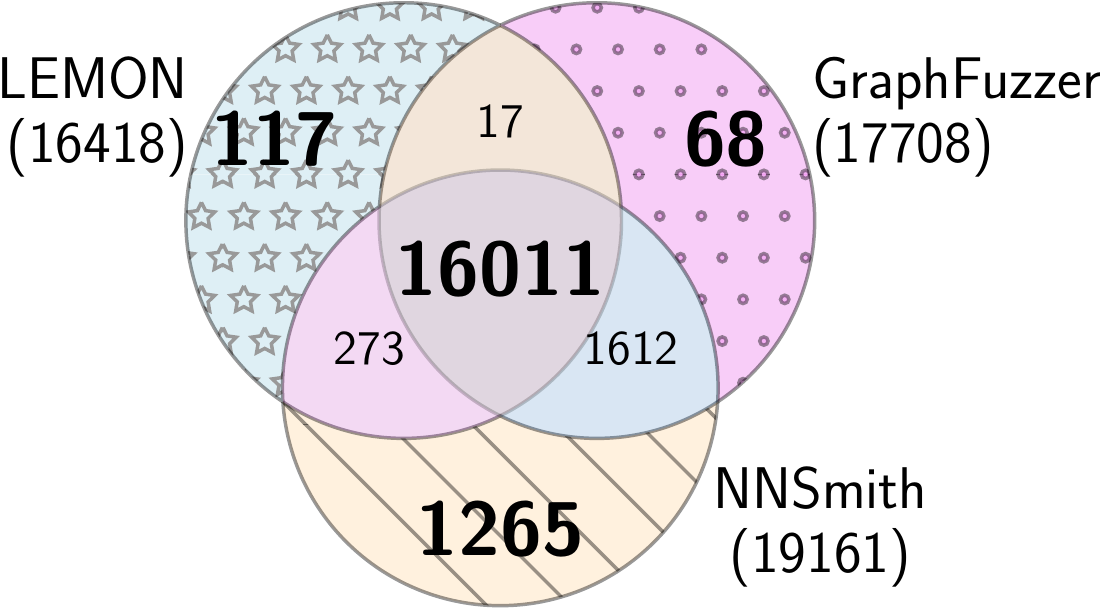}
    \caption{\TVM}
    \label{fig:tvm_venn:all}
\end{subfigure}
\caption{Venn diagram of overall coverage (total coverage shown in parenthesis). %
}
\label{fig:venn}
\end{figure}

We first compare \sys with our graph-level baselines (\ie \GraphFuzz and \LEMON) in terms of code coverage on \TVM and \ORT (since \TensorRT is closed-sourced).
Figure~\ref{fig:totalcov} shows the coverage growth (\emph{y} axis) over four hours (\emph{x} axis).
As is shown in the Figure, \sys beats the 2nd-best baseline (\ie \GraphFuzz) by \covAllHigherFullORT{} on \ORT and by \covAllHigherFullTVM{} on \TVM.
\sys also achieves a decent percentage of total coverage, \ie \covPercORT on \ORT and \covPercTVM on \TVM~\footnote{Note that it is unlikely for \sys to achieve perfect overall coverage as there are many other irrelevant components related to debugging, auto-tuning~\cite{ansor,chen2018learning}, etc.
For instance, existing Linux kernel fuzzers~\cite{kim2020hfl} can only achieve 0.8-10.5\% coverage.}. Figure~\ref{fig:totalcov:iter} further shows the number of generated test cases (\emph{x} axis) within 4 hours and their accumulated total coverage (\emph{y} axis, consistent to Figure~\ref{fig:totalcov}). We can observe that with fewer test cases generated within the same time limit (mainly due to the overhead incurred by constraint solving), \sys can still achieve higher coverage than the 2nd-best baseline (\ie \GraphFuzz), indicating that \sys can generate higher-quality test cases. It is also worth noting that \LEMON is the slowest technique (\eg up to 103$\times$ slower than \sys). The reason is that \LEMON mutates real-world models which can be very costly to run.  
We also have similar observations on the pass-only coverage. For example, as shown in Figure~\ref{fig:passcov}, \sys outperforms \GraphFuzz by \covPassHigherFullORT{} on \ORT and \covPassHigherFullTVM{} on \TVM, showing its effectiveness for testing compiler transformation passes.%

Another interesting observation is that \sys's coverage improvement on \TVM is relatively smaller than that on \ORT (\covAllHigherFullTVM{} v.s. \covAllHigherFullORT{}).
This can be inferred by the difference in their fundamental designs.
While \ORT implements over 130 optimization files targeting various specific graph patterns, \TVM's graph-level optimization is more general.
For example, \TVM's operator fusion does not check specific operator types, but high-level operator properties such as \emph{injective}, \emph{reduce}, etc.
Therefore, \TVM's coverage is less sensitive to the diversity of generated graph patterns.

To show the \emph{unique} coverage for each studied technique, Figure~\ref{fig:venn} further breaks down the coverage sets of different fuzzers through Venn diagrams~\cite{vennwiki}.
It shows that \sys can achieve much higher \emph{unique} coverage than the 2nd-best baseline (\ie \LEMON), \eg \covAllHigherUniORT{} higher on \ORT and \covAllHigherUniTVM{} higher on \TVM.
Despite that \GraphFuzz beats \LEMON in total coverage, \LEMON contrastingly outperforms \GraphFuzz in unique coverage.
This is because \LEMON has a different design from \sys and \GraphFuzz: it mutates existing \emph{real-world} models rather than generating new models from scratch, creating different model patterns. Please note that we omitted the \emph{unique} coverage distribution analysis for pass-only files as it follows a similar pattern as Figure~\ref{fig:venn}.

\begin{figure}[h]
\centering
\begin{subfigure}{0.4\linewidth}
    \includegraphics[width=\linewidth]{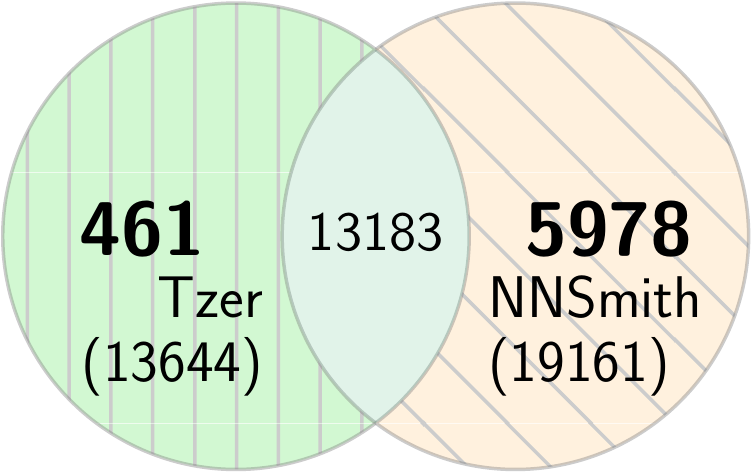}
    \caption{All files.}
    \label{fig:tzer:all}
\end{subfigure}
\hfill
\begin{subfigure}{0.4\linewidth}
    \includegraphics[width=\linewidth]{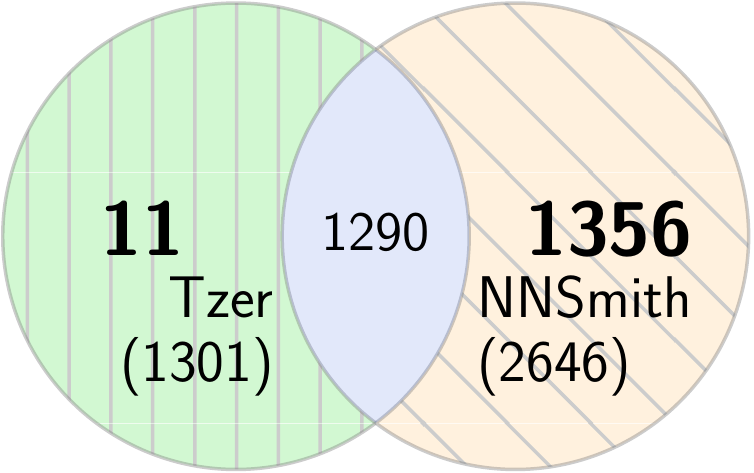}
    \caption{Pass-only files.}
    \label{fig:tzer:pass}
\end{subfigure}
\caption{\sys vs. \Tzer.}
\label{fig:tzer}
\end{figure}

Figure~\ref{fig:tzer} also compares \sys against \Tzer on \TVM (as \Tzer is specifically designed for \TVM). On all \TVM files, \sys as a general graph-level fuzzer, can outperform state-of-the-art IR-level \TVM-specific fuzzer by \covTzerHigher{} in \textit{total} coverage and 13$\times$ in \textit{unique} coverage. %
Interestingly, while other graph-level baselines can at most \textit{exclusively} cover 117
branches (\ie \LEMON in Figure~\ref{fig:tvm_venn:all}), \Tzer has an \textit{unique} coverage of \covTzerUniq{}.
This is because \Tzer directly manipulates low-level IR and some low-level operations are not exposed at the graph level. Moreover, in terms of pass-only coverage, \sys outperforms \Tzer even more, e.g., by 123$\times$ in \emph{unique} coverage, demonstrating the superiority of graph-level fuzzing.

\subsection{Ablation Study}\label{sec:abla}

\begin{figure}[h]
    \centering
    \includegraphics[width=\linewidth]{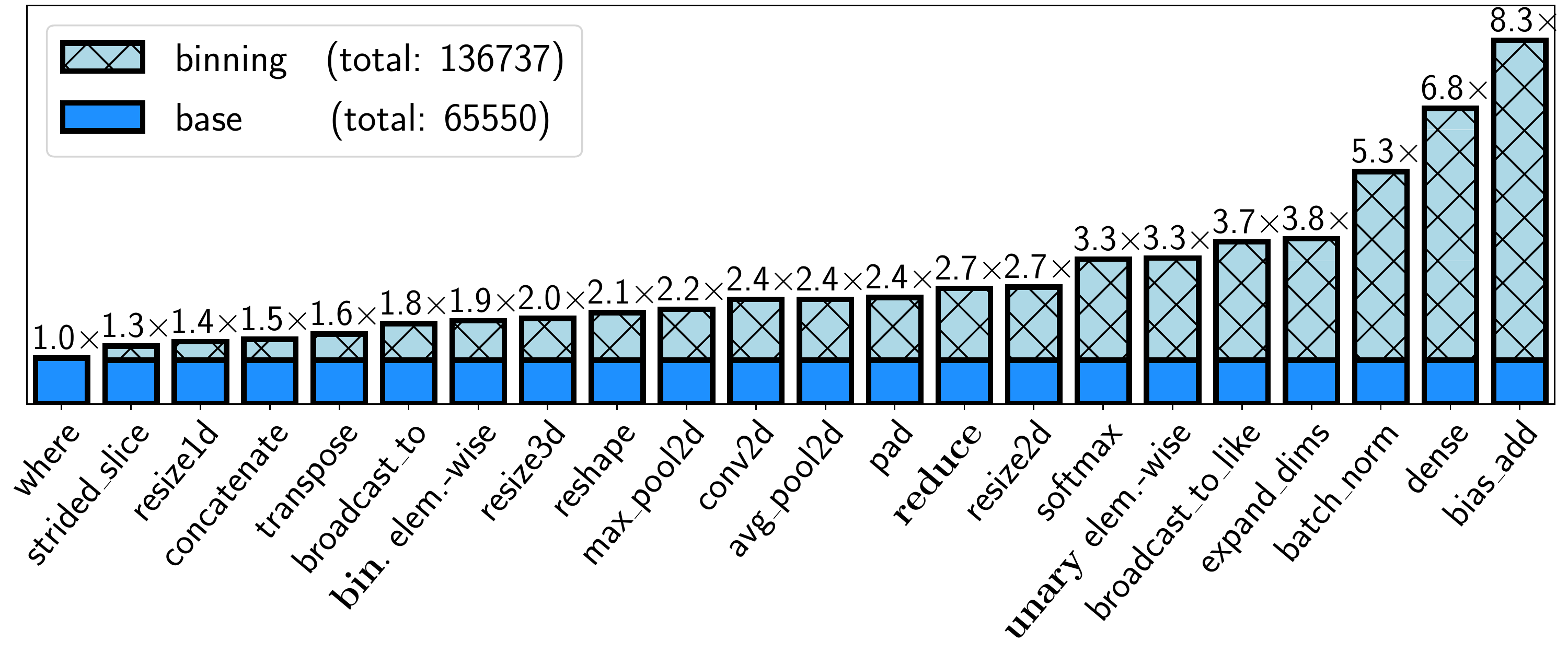}
    \caption{Normalized unique operator instances tested.}
    \label{fig:unique_param_cnt}
\end{figure}

\noindent\textbf{Attribute binning.}\label{sec:eval:binning}
Figure~\ref{fig:unique_param_cnt} evaluates the effectiveness of attribute binning from the perspective of redundancy. Note that for implementation convenience we use the type system from \TVM's Relay IR (parsed from ONNX models) to distinguish operators.%
It shows that within 4 hours, our binning approach achieves \BinOpRatio{} \emph{unique} operator instances, which are distinguished by input types\PageOpt{ (\ie shapes and data types)} and operator attributes.

\begin{figure}[h]
\centering
\begin{subfigure}{0.4\linewidth}
    \includegraphics[width=\linewidth]{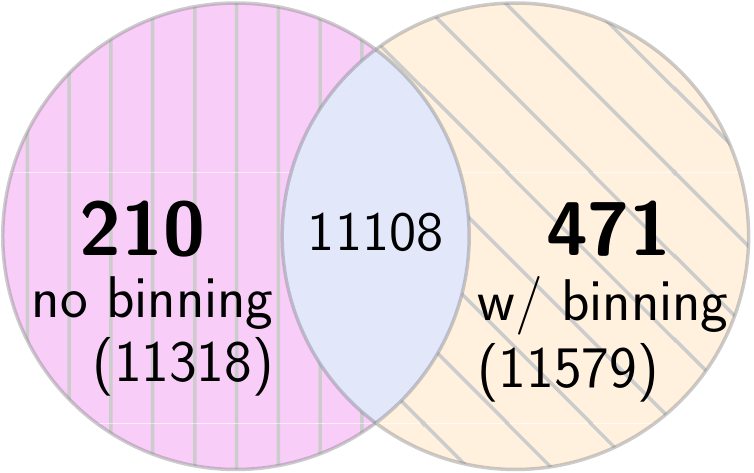}
    \caption{\ORT}
    \label{fig:binvenn:ort}
\end{subfigure}
\hfill
\begin{subfigure}{0.4\linewidth}
    \includegraphics[width=\linewidth]{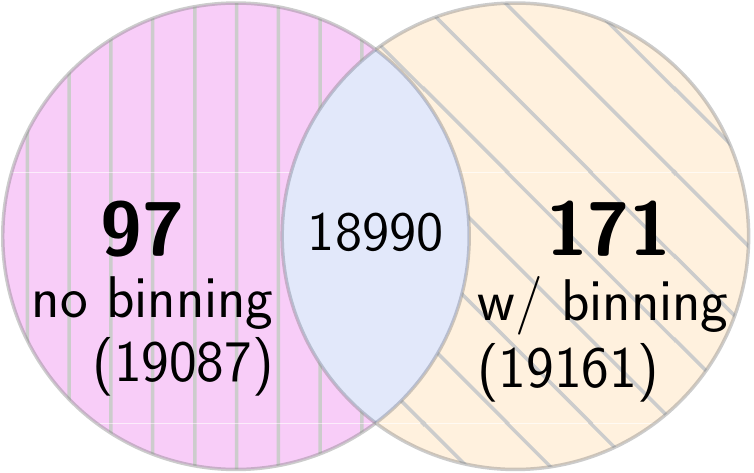}
    \caption{\TVM}
    \label{fig:binvenn:tvm}
\end{subfigure}
\caption{Impact of attribute binning on coverage.}
\label{fig:binvenn}
\end{figure}

Turning to system coverage, as shown in Figure~\ref{fig:binvenn}, attribute binning improves the \textit{unique} branch coverage by \BinUniqRatioORT{}
for \ORT (Figure~\ref{fig:binvenn:ort}) and \BinUniqRatioTVM{}
for \TVM (Figure~\ref{fig:binvenn:tvm}).
The total coverage improvement is relatively subtle (up to \BinTotalRatioORT{}) as the binning approach aims at covering the hard-to-hit branches whose proportion is expected to be minor. %
For example, simply importing \TVM's libraries with ``\texttt{import tvm}'' can hit 4015 branches but those branches are unlikely to have bugs.%

\begin{figure}
    \centering
    \includegraphics[width=\linewidth]{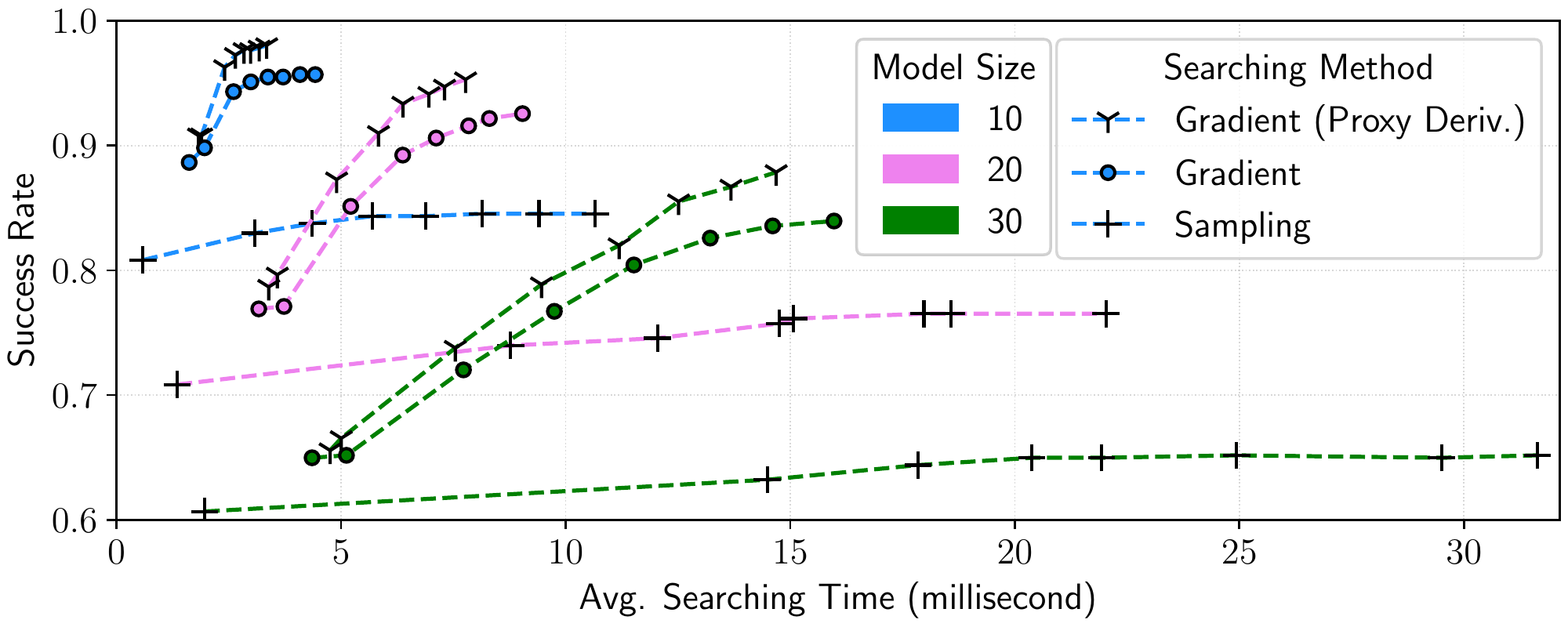}
    \caption{Effectiveness of gradient-based search.}
    \label{fig:grad}
\end{figure}

\noindent\textbf{Gradient guidance.}\label{sec:eval:grad}
Figure~\ref{fig:grad} evaluates the effectiveness of three input/weight searching methods:
1) {\fontfamily{lmss}\selectfont Sampling}: randomly initializing test case values; 2) {\fontfamily{lmss}\selectfont Gradient (Proxy Deriv.)}: searching values via the full gradient-based approach; and 3) {\fontfamily{lmss}\selectfont Gradient}: method two without proxy derivatives.
The experiment is conducted on three model groups, each of which contains 512 models of 10, 20 and 30 nodes respectively. Every model has at least one vulnerable operator.
The \emph{Sampling} baseline randomly samples values from the range of $[1, 9]$ which is empirically obtained selecting the best one from various tested ranges. %
For fairness, all methods run on the same groups of models with the same initial weights/inputs generated by the \emph{Sampling} baseline.
We assign different per-model searching timeouts (\ie $i\times 8$ms where $i\in[1,8]$) to each method and observe the ratio of models with numeric-valid inputs/weights (\emph{y}-axis) over group-wide average searching time (\emph{x}-axis).
Figure~\ref{fig:grad} shows that our full gradient search improves the numerical validity of \emph{Sampling} by 1.16-1.34$\times$ as the node size/difficulty grows. Also, the proxy derivative mechanism consistently helps our gradient search achieve higher success rate within shorter amount of time.%

We also observe that searching time is negligible compared with model generation time, \eg generating a 10-node model costs 83ms on average while our gradient-based searching only takes 3.5ms (\textbf{4.2\%}) to achieve a success rate of \textbf{98\%}.

\subsection{Bug Study}\label{sec:eval:bug}

\begin{table}[h]
\adjustbox{max width=\linewidth}{
\newcolumntype{M}[1]{>{\hspace{0pt}}m{#1}}
    \centering
\begin{tabular}{ccccc}
        \hline
        & Transformation & Conversion & Unclassified & Total      \\
        \hline
\ORT              & 10        & 0      & 2   & 12    \\
\TVM              & 29        & 11     & 0   & 40   \\
\TensorRT         & 4         & 2      & 4   & 10    \\
\PyTorch Exporter & --        & 10     & --  & 10    \\
        \hline
\makecell{\textbf{Total}\\(Crash/Semantic)}   & \makecell{43 \\(34/9)}      & \makecell{23\\ (18/5)}  & \makecell{6\\ (3/3)} & \makecell{72\\ (55/17)} \\
        \hline
\end{tabular}
    }
    \caption{Bug distribution.}
    \label{tab:bug-breakdown}
\end{table}

To date, \sys has uncovered \textbf{\bugTotal} \emph{new} bugs as shown in Table~\ref{tab:bug-breakdown}, where \textbf{\bugConfirmed} have been \emph{confirmed} and \textbf{\bugFixed} have been \emph{fixed}.
Others are awaiting developer responses.
Interestingly, in addition to compiler bugs, since \sys generates models through \PyTorch ONNX exporter (\S\ref{sec:impl}), it also found \bugTorch conversion bugs in \PyTorch as a by-product.
Among the bugs we found, \bugIncon{} are semantic bugs (result inconsistencies with \PyTorch) and \bugCrash{} are crash bugs (segmentation faults or exceptions).
In total, there are \textbf{\bugTrn} \textit{transformation} bugs in \ORT (\bugTrnORT), \TVM (\bugTrnTVM) and \TensorRT (\bugTrnTRT), accounting for the majority of the detected bugs \footnote{We classify bugs first based on code inspection (when possible); otherwise, we classify a bug as transformation bug if its individual operators cannot reproduce the issue separately.}. %
We found that most of these were optimization bugs: of the \bugTrnFixed fixed transformation bugs we found, \textbf{\bugOptFixed} are \textit{optimization} bugs (and the remaining one is an \emph{unclassified} bug in \TensorRT whose code is not available).

Of the \bugTotal{} bugs we found, \textbf{\bugOurOnly{}} bugs cannot be triggered using the algorithms implemented by \LEMON or \GraphFuzz. Of these \bugOurOnlyTrn{} are transformation bugs and \bugOurOnlyCvt{} are conversion bugs. \LEMON's algorithms can trigger at most  \bugLEMON of all bugs we found, while \GraphFuzz's algorithms can trigger at most \bugGraphFuzz of these.
The core difference is that these prior approaches limit how non-shape preserving operators are connected in the graph, thus limiting graph diversity.
In addition to this theoretical analysis, we also evaluated all tools by running them for four hours under the \emph{same} setting (\eg all on the \emph{default} compiler versions as shown in \S~\ref{sec:setup}),
\sys triggers \textbf{38} \textit{unique} crashes (by error messages) for \ORT and \textbf{13} for \TVM, while \LEMON triggers none and \GraphFuzz only triggers 1 crash for each of \ORT and \TVM.
For instance, the only \ORT bug detected by \GraphFuzz is the wrong fusion to a double-precision \emph{ReLU-Clip} connection (element-wise and thus shape-preserving).

We next describe transformation and conversion bugs we found by illustrating prominent bug patterns with examples. We use \markStar
 to denote bugs \emph{exclusively} found by \sys.

\noindent\textbf{Transformation bugs.}
    \emph{Wrong expression simplification:}
    We found \bugSimpl such bugs in \ORT (\bugSimplORT) and \TVM (\bugSimplTVM).
    One bug\markOur{} happens in \texttt{FuseMatMulScale} when \ORT optimizes $(s_a\cdot A)@(s_b\cdot B)$ to $(s_a\cdot s_b)\cdot(A@B)$ for scalars $s_a, s_b$ and matrices $A,B$ where $@$ denotes \emph{MatMul}. However, when $B$ is a $1\times 1$ matrix, \ORT can mistake matrix $B$ as a scalar and rewrite it into $(s_a\cdot B)\cdot(\underline{A@s_b})$, which is \underline{illegal} as \emph{MatMul} does not accept scalar inputs, causing a compiler exception.
    Prior work cannot use the non-shape-preserving \emph{MatMul} operator%
    , thus missing such bugs.
     Wrong expression simplification can also lead to \emph{semantic bugs}, which may lead to wrong decisions in downstream AI applications, introducing security threats in critical scenarios (\eg self-driving).
    For example, \TVM has a buggy arithmetic optimization pass that switches the order of division and multiplication when rewriting $\lfloor\frac{x\;\mathrm{mod}\;y}{i}\rfloor \times i\;\mathrm{mod}\;z$, simplifying it to $(x\;\mathrm{mod}\;y)\;\mathrm{mod}\;z$ incorrectly.
    
\emph{Wrong layout analysis:}
        Memory layout optimizations in \TVM first rewrite layouts of the most beneficial operators (\eg \emph{Conv2d}) to efficient ones\PageOpt{ (\eg for vectorization)} and then let remaining operators adapt changed layouts.
        We found \bugTrnTVMLayoutExcl layout transformation bugs\markOur in \TVM, related to non-shape-preserving operators including broadcasting, reduce and slicing, which cannot be handled by prior work. %
        For example, \TVM can rewrite N\underline{C}HW \emph{Conv2d} to the SIMD-friendly N\underline{$\frac{C}{4}$}HW\underline{4c} layout (NCHW4c for short), by packing every 4 elements on \underline{C} to the new sub-dimension (\underline{4c}).
        However, using this optimization when the \emph{Conv2d} is followed by a \emph{Slice} operator whose stride for \underline{C} is greater than one %
        causes \TVM to crash. %
        \GraphFuzz cannot find this bug because to ensure shape alignment it always uses a stride of 1.%

\emph{Integer type mismatch:} 
        DL compilers, like traditional compilers (\eg LLVM~\cite{lattner2002llvm}), leverage IRs to simplify optimization.
        IR type mismatch can happen if one pass makes wrong assumption for the IR being transformed.
        This is especially a pain for \TVM:
        we found \bugIntMismatch bugs\markOur stopping the compilation due to \texttt{int32}-\texttt{int64} mismatch and one core \TVM developer also admits that \emph{``\TVM has a pretty fragile system of using i32 vs i64; I personally experienced it a few times before...''.}%
        \texttt{int64} is often introduced by shape-related operators (\eg shape attributes of \emph{Reshape} and \emph{BroadcastTo}),
        which are not supported by prior work as they cannot handle those complicated shape constraints.
        Since our first bug report on such issues, there have been 12 fixes (7 from us and \textbf{5} from followers)
        within 5 months to resolve similar issues, one of which even blocked models in production.
        Interestingly, a bug we found also helped the developers find another bug that had previously
        been diagnosed as the outcome of a flaky test~\cite{luo2014empirical}.

\PageOpt{\sys also finds \bugConv conversion bugs where the \PyTorch ONNX Exporter exports ill-formed models or DL compilers parse the incoming model in a wrong way.
These bugs are mainly due to \emph{unhandled edge cases of scalar tensors} (\bugConvScalar), \emph{wrong broadcasting} (\bugConvBcast), or \emph{data type mismatch} (\bugConvType). }
\noindent\textbf{Conversion bugs.}
\emph{Wrong scalar handling}:  %
        We found 6 crash bugs\markOur triggered when \TVM imports \emph{reduce}-like operators with a scalar input.
        Since these operators are not shape-preserving, prior work cannot trigger such bugs.
        Similarly in \PyTorch, when exporting \emph{Log2} with a scalar input, the exporter mistakenly sets its output to a rank-1 tensor instead of a scalar, causing a \textit{semantic} issue.
        A few days after our report, developers identified \textbf{37} other \emph{similar bugs}. Concurrently, \sys also identified a subset of these bugs, but in our evaluation we only treat the first bug (\emph{Log2}) as one found by \sys.

\emph{Wrong broadcasting:}
        Given a 3-way broadcasting operation\linebreak \emph{Where}$(C_{1\times 1},T_{3\times 1},F_2)$, a \TVM bug\markOur{} causes the lower-ranked tensor $F_2$ being ignored during shape inference, resulting in the wrongly inferred shape $3\times 1$, which should be $3\times 2$.
        This incurs a compiler failure in later phases.
        Another \TVM bug\markOur{} causes an import failure to \emph{MatMul} with single-rank broadcasting (one input is a vector) and notably, one month after our bug report, real-world \TVM users also encountered such issues and pushed for its fix, showing that \sys can synthesize real-world model patterns.
        Prior work cannot detect them since their design are incompatible with broadcasting operations.

 \emph{Data type mismatch:}
 Operators' data type supports vary by ONNX versions, which are often mishandled. For example, \PyTorch can mistakenly (and silently) export \emph{Clip} whose data type is \texttt{int32} which is not supported by ONNX version 11.
Such ill-formed models will be rejected by most compilers; however, it can also be mistakenly compiled by \TensorRT, producing unexpected model outputs (\ie semantic bugs in \TensorRT), due to the wrongly interpreted attributes.

\noindent\textbf{False alarms.} As we discussed in the introduction, floating point semantics~\cite{Priest92onproperties, overton2001numerical} mean that even correct optimizations can lead to scenarios where an optimized model's output differs from the reference output. Consequently, we check output equivalence by checking that the distance between model outputs,when scaled by their overall magnitude is small. However, in some cases valid optimizations can lead to a large relative change in outputs and produce false alarms. For example, when feeding the \emph{Sigmoid} with a large value, the optimized output can be 1 whereas the reference one is less than (though very close to) $1$. 
When it provides the input to a \emph{Floor} operator, the optimized output will differ from the reference output by $1$,
To reduce false alarms, we use a high error tolerance in comparison. In addition, false positives share similar structural patterns (\eg \emph{Sigmoid} followed by \emph{Floor} and \emph{Equal} for float-pointing inputs) that we can easily filter. Thus it did not cause large trouble for us.

%% file: sections/6-related-wk.tex
\section{Related Work}\label{sec:rw}
Since the first proposal of fuzzing~\cite{miller1990empirical}, various techniques have been proposed for fuzzing systems of different application domains~\cite{boehme2021fuzzing, li2018fuzzing, manes2019art, zou2021tcp, trippel2021fuzzing, dewey2014language, c11tester, pmfuzz, neal2021hippocrates, dce}. In this section, we mainly talk about the most closely related work in \dl system fuzzing and compiler fuzzing.

\subsection{\dl System Fuzzing}
\label{sec:rwmlsystest}

In recent years, a number of techniques have been proposed to test \dl libraries and compilers. As one of the first techniques in this direction, \cradle~\cite{marcozzi2019compiler} directly runs existing DNN models on different \dl libraries to detect potential inconsistencies via differential testing.%
Later on, \audee~\cite{guo2020audee} and \LEMON~\cite{lemon} further extend \cradle by applying search-based mutation strategies on the DNN models and their inputs to cover more library code.
While \audee mainly focuses on mutating layer parameters and weight/input tensors, \LEMON further applies more advanced mutation rules, including layer deletions/additions. Meanwhile, to ensure correctness of generated models, \LEMON~\cite{lemon} only mutates type-preserving operators (or blocks of operators) from the real-world models, to avoid handling type constraints.%
However, there are many non-shape-preserving operator types, \eg even the commonly used \emph{Conv2d} cannot be completely handled by \LEMON.
More recently, \GraphFuzz~\cite{luo2021graph} allows a slightly larger operator search space using padding/slicing to align unmatched tensor shapes and also specifically controls the attributes of shape-changing operator types to create shape-preserving instances (\eg \emph{Conv2d} with kernel size/stride of 1). However, this design still substantially limits model diversity (as demonstrated in \S\ref{subsec:bug-ex}).
The very recent (and concurrent) \muffin work~\cite{muffin} shares a similar limitation as \GraphFuzz: it uses ``reshaping'' layers to align tensor shapes during model generation; in addition, \muffin focuses on finding bugs in traditional DL libraries rather than DL compiler bugs. In this work, we aim to support more diverse/valid model generation for DL compiler fuzzing via a fundamentally different design powered by symbolic constraint solving~\cite{cadar2013symbolic} and gradient-driven search.%

To complete \dl system testing at the model level, researchers have also proposed \dl system fuzzing techniques focusing on directly generating or manipulating the low-level model IRs~\cite{tvmfuzz, tzer}. \tvmfuzz~\cite{tvmfuzz} aims to automatically generate arbitrary low-level IRs based on a set of predefined grammar rules for fuzzing the popular TVM compiler~\cite{tvm}. 
The more recent \Tzer work~\cite{tzer} leverages coverage feedback to perform joint mutation of both the low-level IR and optimization passes for TVM.
While \Tzer has shown promising results over \tvmfuzz, the low-level IR mutation adopted by \Tzer can hardly test the graph-level optimizations widely adopted by various \dl compilers (as shown in \S\ref{sec:eval:cov}).   

In recent years, researchers have also investigated techniques to fuzz each \dl library API in isolation.
Since \dl APIs are usually exposed in Python, a dynamically typed language, prior techniques, such as \predoo~\cite{zhang2021predoo}, require users to manually set up the function arguments, and can only be evaluated on a limited number of APIs.
To address this challenge, \FreeFuzz~\cite{wei2022free} dynamically traces API executions from various sources (including documents, developer tests, and model zoos), and generates new tests by mutating  traced inputs. 
More recently, DeepREL~\cite{deeprel} aims to automatically infer relational APIs (e.g., APIs that return the same values/statuses when given the same inputs), and then leverages such API relations as test oracle for catching more bugs than \FreeFuzz.%
While such API-level testing techniques are adequate for testing first-generation \dl libraries (\S\ref{sec:rwdnn}), they can hardly find bugs in graph-level optimizations (e.g., 86\% of the transformation bugs detected by \sys require multiple operators to trigger).

\subsection{Compiler Fuzzing}

As one of the most widely studied compiler fuzzing approaches in the literature~\cite{marcozzi2019compiler}, \emph{grammar-based} techniques (such as \csmith~\cite{yang2011finding}, \jsfun~\cite{jsfunfuzz}, and \langfuzz~\cite{holler2012fuzzing}) aim to generate syntactically valid input programs acceptable by the underlying compilers. 
While effective, it is hard for grammar-based techniques to ensure the semantic correctness of the generated programs to cover deep code paths, and highly specialized analyses have to be employed for specific languages. Therefore, various \emph{mutation-based} techniques~\cite{le2014compiler, donaldson2017automated, sun2016finding, le2015finding, zhang2017skeletal} have also been proposed for fuzzing compilers via mutating existing seed input programs.%
Moreover, given the advances in \dl, researchers have also proposed \emph{learning-based} techniques for compiler fuzzing. 
\deepsmith~\cite{cummins2018compiler} and \deepfuzz~\cite{liu2019deepfuzz} directly leverage recurrent neural networks (RNNs) to generate test programs from scratch, while \montage~\cite{lee2020montage} performs mutation-based fuzzing, and replaces code snippets of the seed programs with new code fragments generated by RNNs.
More recently, researchers have also leveraged the advanced pre-trained language models (e.g., \gpt~\cite{brown2020language}) for more powerful test program generation for compiler fuzzing~\cite{ye2021automated}.
Such existing compiler fuzzing techniques can be potentially applied to the low-level IRs (C-like) for fuzzing \dl compilers~\cite{tzer}. However, they can be hardly directly applied for graph-level \dl compiler fuzzing, and our study has also shown the superiority of \sys over state-of-the-art IR-level DL compiler fuzzer. %

%% file: sections/7-conclusion.tex
\section{Conclusion}
\sys is a tool for generating diverse and valid test cases for deep learning compilers. It creates abstract operator models to ensure the validity of the generated models, and further utilizes incremental graph generation and attribute binning to ensure its diversity. To avoid false alarms and bug escapes, \sys leverages gradient search to find inputs that do not introduce NaN/Inf in the computation. \sys is easily extensible to support new operators with few lines of code.  Lastly, \sys is implemented to generate models in the popular format ONNX and is readily applicable to any systems with ONNX support. To date \sys has found \bugTotal new bugs in \TVM, \TensorRT, \ORT, and PyTorch, \bugConfirmed of which have been confirmed or fixed, demonstrating its effectiveness.

\section*{Acknowledgments}
We thank the ASPLOS reviewers for their insightful comments. We also thank Yuanyi Zhong, Lingfan Yu and Leyuan Wang for insightful discussions in the early stages of the project, Jinjun Peng for helping open-source the project, and the the NYU IT High Performance Computing group for providing computing resources. This work was partially supported by the National Science Foundation grants CCF-2131943 and CCF-2141474, a Google research award, a Meta research award, an AMD research award, and a gift from Microsoft Corporation.

%% file: ae.tex
\appendix
\section{Artifact Appendix}

\subsection{Abstract}

The artifact contains evidence of bug finding, source code of \sys's prototype, and user-friendly HTML documentation for re-generating the results.
Specifically, it includes (1) links to bugs reported by the authors as real-world bug finding evidence, and (2) scripts and code to re-generate main results in \S~\ref{sec:eval}.
To make artifact evaluation as simple as possible, our artifact is packaged into a pre-built docker image, along with a detailed and friendly HTML documentation.
To fully evaluate the artifact, a X86-CPU platform with docker access is needed, with approximately 21 hours of machine time and 1 hour of manual inspection time.

\subsection{Artifact check-list (meta-information)}

{\small
\begin{itemize}
  \item {\bf Run-time environment: } Linux.
  \item {\bf Hardware: } X86 CPU.
  \item {\bf Metrics: } C++ source-level branch coverage.
  \item {\bf How much disk space required (approximately)?: } 256GB.
  \item {\bf How much time is needed to prepare workflow (approximately)?: } A few minutes to install and import the docker container image.
  \item {\bf How much time is needed to complete experiments (approximately)?: } About 21 hours of machine time and 1 hour of manual inspection time.
  \item {\bf Publicly available?: } Yes.
  \item {\bf Code licenses (if publicly available)?: } Apache-2.
  \item {\bf Archived (provide DOI)?: } 10.5281/zenodo.7222132
\end{itemize}
}

\subsection{Description}

\subsubsection{How to access.}

The artifact can be downloaded from \url{https://zenodo.org/record/7222132}.

\subsubsection{Hardware dependencies.}

A computer with X86 CPU and disk space of over 256 gigabytes.

\subsubsection{Software dependencies.}

Docker.

\subsection{Installation}

There are two common ways to install the docker image:

To directly install the docker image from the Docker Hub (\url{https://hub.docker.com/repository/docker/ganler/nnsmith-asplos23-ae})\linebreak repository: \texttt{docker pull ganler/nnsmith-asplos23-ae}

To import the image locally from the Zenodo repository (\url{https://zenodo.org/record/7222132}):

\texttt{tar xf NNSmith-ASPLOS23-Artifact.tar.gz}

\texttt{export IMG=ganler/nnsmith-asplos23-ae:latest}

\texttt{cat nnsmith-ae.tar | docker import - \$IMG}

\subsection{Experiment workflow}

Please open \texttt{html/index.html} in your browser or use the online documentation (\url{http://nnsmith-asplos.rtfd.io/}) for detailed installation and evaluation steps.

The overall logical workflow of experiments includes:

\begin{enumerate}
    \item Install and run the docker image.
    \item Running \sys and baselines on instrumented \TVM and \ORT for 4 hours each to get code coverage reports.
    \item Visualize coverage in coverage trends and Venn diagram.
\end{enumerate}

\subsection{Evaluation and expected results}

All instructions and steps to reproduce our experimental results are in the artifact documentation (\texttt{html/index.html} or \url{http://nnsmith-asplos.rtfd.io/}).

The steps for main experiments are included in the ``Evaluating artifact'' chapter which re-generates results in \S~\ref{sec:eval:cov} and \S~\ref{sec:eval:bug}.
We also include instructions to re-generate non-main experiments (\ie ablation study in \S~\ref{sec:abla}) in the ``Read more'' chapter.